\pdfoutput=1

\documentclass[11pt]{article}

\usepackage{acl}

\usepackage{times}
\usepackage{latexsym}

\usepackage[T1]{fontenc}

\usepackage[utf8]{inputenc}

\usepackage{float}
\usepackage{inconsolata}
\usepackage{microtype}
\usepackage{amsmath}
\usepackage{amssymb}
\usepackage{graphicx}

\usepackage{booktabs}
\usepackage{multirow,multicol}
\usepackage{microtype}

\usepackage[linesnumbered,ruled,vlined]{algorithm2e}
\usepackage{amsfonts}
\usepackage{tabularx}
\usepackage{subfigure}
\usepackage{layouts}
\usepackage{cleveref}
\usepackage{todonotes}
\usepackage{xcolor}
\usepackage{soul}

\usepackage{pifont}
\newcommand{\cmark}{\ding{51}}
\newcommand{\xmark}{\ding{55}}

\newcommand{\taskfullname}{Multimodal Review Helpfulness Prediction}
\newcommand{\taskabbrname}{MRHP}
\newcommand{\modelname}{PREMISE}
\newcommand{\modelfullname}{PREdict with MatchIng ScorEs}
\newcommand{\csfigheight}{1.1cm}

\title{PREMISE: Matching-based Prediction for Accurate Review Recommendation}

\author{Wei Han$^\dagger$, Hui Chen$^\clubsuit$
\thanks{Corresponding to hui.chen@nus.edu.sg}
, Soujanya Poria$^\dagger$,\\
  $^\dagger$ Singapore University of Technology and Design, Singapore\\
  $^\clubsuit$ National University of Singapore, Singapore \\
}

\begin{document}
\crefformat{section}{\S#2#1#3}
\crefformat{subsection}{\S#2#1#3}
\crefname{algocf}{alg.}{algs.}
\Crefname{algocf}{Algorithm}{Algorithms}

\maketitle
\begin{abstract}
We present~\modelname~(\modelfullname), a new architecture for the \textit{matching-based} learning in the multimodal fields for the~\taskfullname~(\taskabbrname) task. 
Distinct to previous fusion-based methods which obtains multimodal representations via cross-modal attention for downstream tasks, 
\modelname~computes the multi-scale and multi-field representations, filters duplicated semantics, and then obtained a set of matching scores as feature vectors for the downstream recommendation task.
This new architecture significantly boosts the performance for such multimodal tasks whose context matching content are highly correlated to the targets of that task, compared to the state-of-the-art fusion-based methods.
Experimental results on two publicly available datasets show that~\modelname~achieves promising performance with less computational cost.
\end{abstract}
\section{Introduction}
The e-commerce industry has experienced an unprecedented boom in the past decade.
Powered by an instant trading system, online shopping platforms successfully endow buyers who seek their favorite goods and sellers who advertise their products with convenience for transaction~\citep{boysen2019warehousing,vulkan2020economics,alfonso2021commerce}.
However, when wandering through these shops, customers easily fall into the dilemma of deciding whether to buy a product displayed on the screen.
At that time, the comments left by past customers are often considered as the most valuable reference.
Therefore, how to automatically evaluate review's quality and accurately recommend these reviews becomes a challenge yet an opportunity for online shopping platforms to attract and hold customers.
Formally, researchers formulate this problem as the Review Helpfulness Prediction (RHP) task~\citep{tang2013context,ngo2014influence}, which aims to quantify the value of each review to potential customers.
By sorting these reviews according to the predicted helpfulness scores in descending order, the platform can post the most valuable reviews at the conspicuous location in the shop page.
\begin{table}[t!]
    \centering
    \small
    \resizebox{\linewidth}{!}{
        \begin{tabular}{p{\linewidth}}
        \toprule
        \textbf{Product Name}: Gourmia GK250 (1.8 Qt/1.7 L) Cordless Stainless Steel Kettle Supreme - Speed Boil - Auto Shutoff Boil Detect - Concealed Element - 360 Swivel Base - 1500 Watts \\
        \midrule
        \textbf{Product description}: Forget fumbling with all of those microwaves...Heat up to 1.7 liters of piping hot water in a flash! ... and auto shut off to ensure completely safe kettle usage...
        
        \subfigure{\includegraphics[height=\csfigheight]{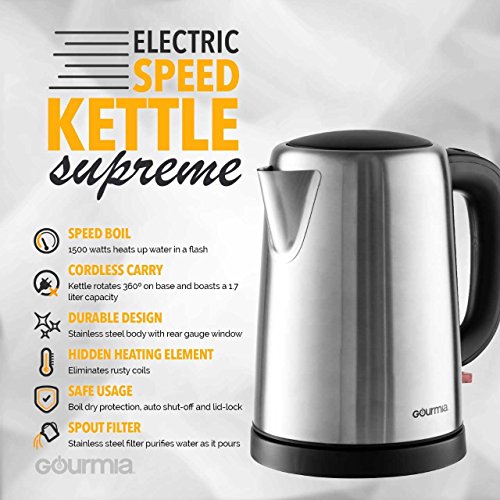}}
        \subfigure{\includegraphics[height=\csfigheight]{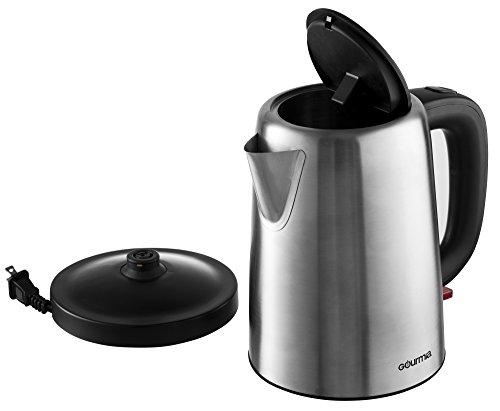}}
        \subfigure{\includegraphics[height=\csfigheight]{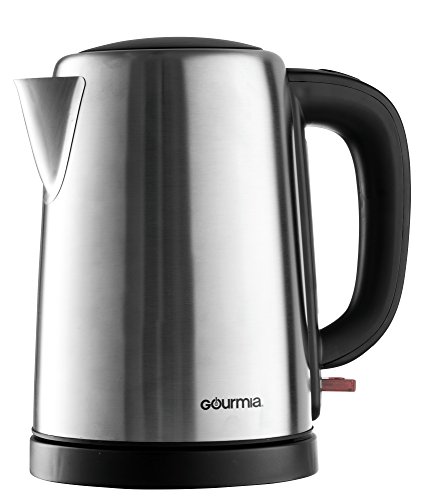}}
        \subfigure{\includegraphics[height=\csfigheight]{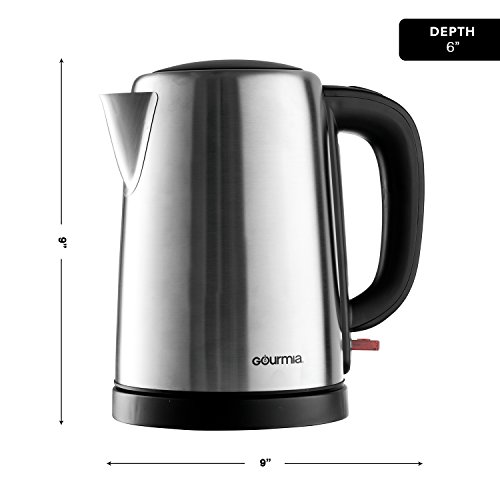}}
        \subfigure{\includegraphics[height=\csfigheight]{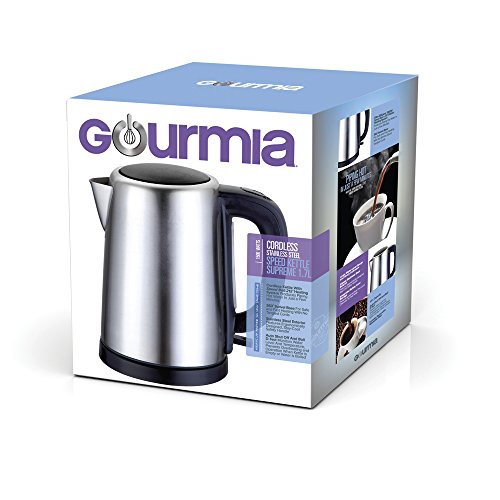}}
        \subfigure{\includegraphics[height=\csfigheight]{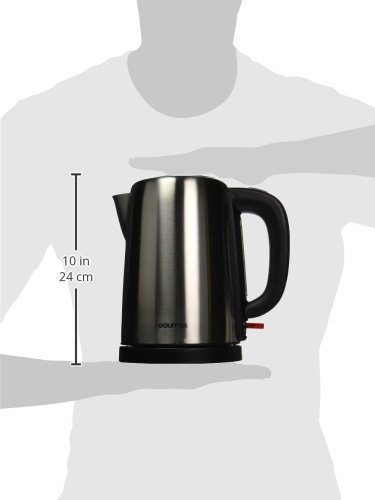}} \\
        \midrule
        \textbf{Review 1 (Helpfulness Score: 4)}: I've had my eye on an eighty dollar stainless steel electric kettle for a while, but I didn't care to make that investment just yet. 
        ...
        Kettle automatically turned off within seconds of reaching a full boil. \textbf{When filled halfway (to 1 liter), it takes about 4 and a half minutes to boil.}
        \textbf{This kettle doesn't take up much counter space either, it's easy to tuck in to the corner by my stove when I need it out of the way.} Overall I'm pretty happy with this, and \textbf{thankful for a kettle that turns itself off so I don't have to worry about forgetting it while it's boiling.}

        \subfigure{\includegraphics[height=\csfigheight]{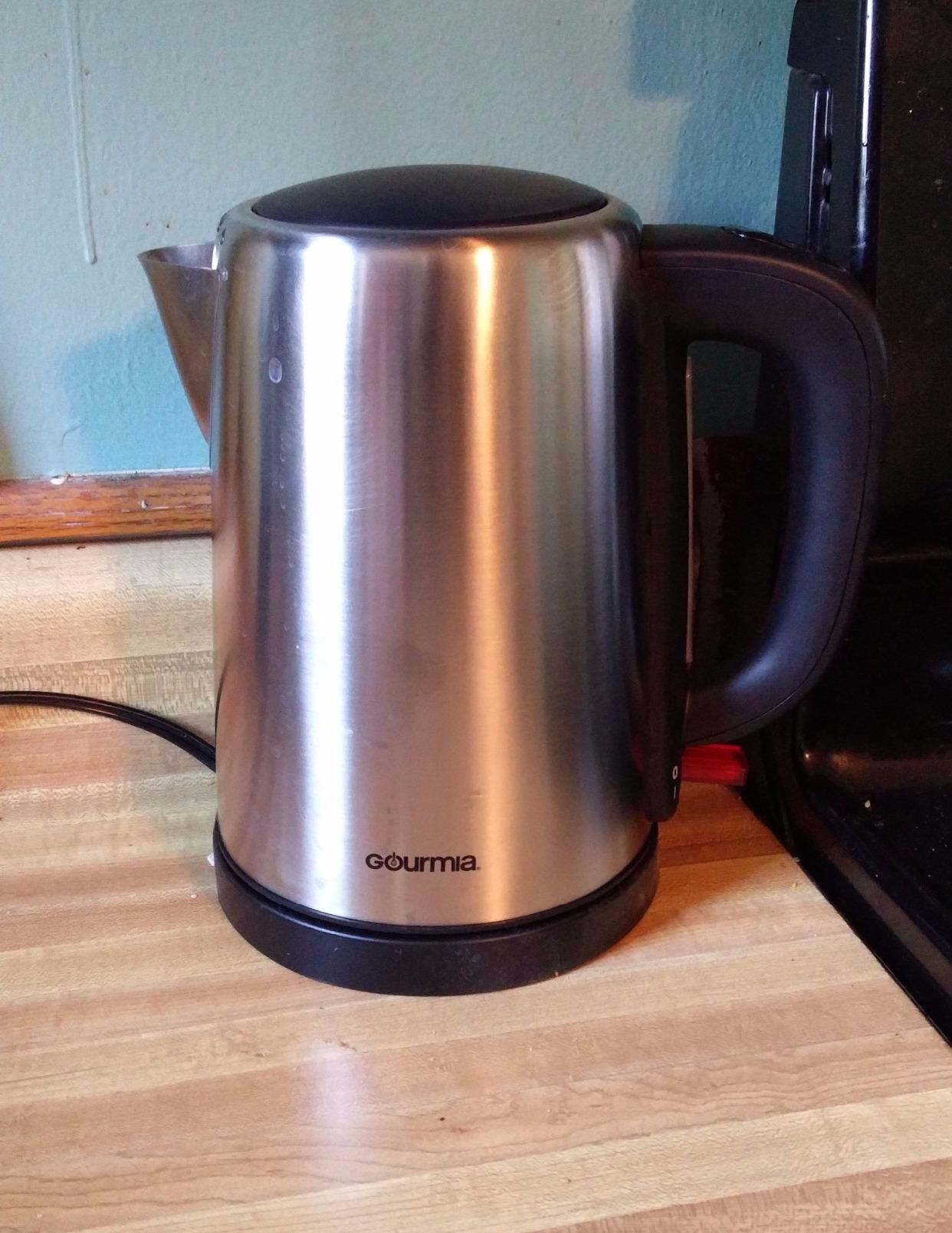}} \\
        \midrule
        \textbf{Review 2 (Helpfulness Score: 1)}: The handle for the kettle which love or I thought I did is falling off.  What do I do?  Was this product meant to only last for a few months? I need help here!  So disappointing.  I did not want to report it here, but I see nowhere else to do so.
        
        \subfigure{\includegraphics[height=\csfigheight]{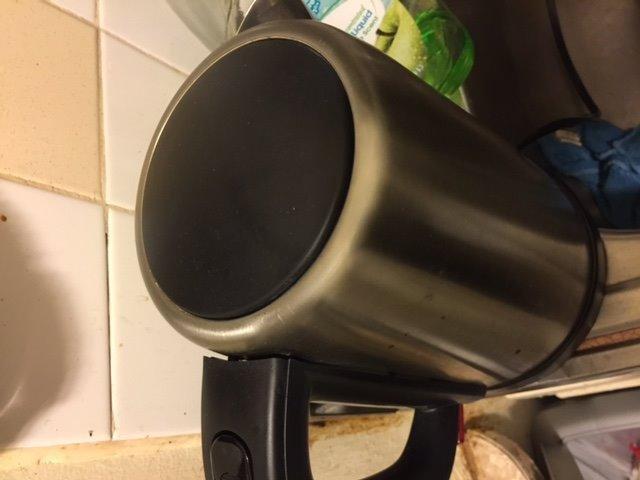}} \\
        \bottomrule
        \end{tabular}
    }
    \caption{A pair of reviews with high and low helpfulness scores from Amazon-MRHP dataset. 
    We highlight the text that provides customers helpful information. 
    Due to space limitation we only preserve key sentences in the product description.}
    \label{Problem Setting}
\end{table}
In recent years, RHP task has been extended to the multimodal scenario by incorporating the text-attached images as an auxiliary source to help the model make more accurate predictions~\citep{liu2021multi}, termed Multimodal RHP (MRHP).

Previous achievements to address this problem
usually employ fusion modules to learn expressive multimodal representations for prediction~\cite{arevalo2017gated,chen2018cross,liu2021multi,han-etal-2022-sancl,nguyen-etal-2022-adaptive}.
Despite the gained satisfying results, there are still several drawbacks in those models that limit the system's performance.
First, explicit multi-scale modeling is missing.
Extant works usually take into account single or combinations of segments in a \textit{fixed scale}, such as tokens, phrases and image patches.
Nevertheless, multi-scale modeling is necessary especially when confronted with long textual inputs like reviews, 
since it has been pointed out that task-related information is commonly distributed unevenly among all the sentences~\cite{chen2019multi}.
Take a randomly picked product and attached review in Table~\ref{Problem Setting}~as an example, even a review of high helpfulness score (Review 1), there are many dispensable sentences (unbold text) that stray from the product it comments on (off-the-topic).
Second, though fusion-based models have been demonstrated effective in a family of multimodal tasks, they often result in bulk structures and hence are time-consuming in the training process~\cite{nagrani2021attention}. 
Previous research reveals that semantic matching, i.e., the similarity between semantic elements 
(image regions, text tokens and their n-grams) 
can be regarded as a crucial factor that guide models to make the final decision~\cite{ma2015multimodal,huang2017instance,liu2017learning}.
Based on the discovery, we postulate that quantified matching scores could be fully exploited in regression.  
Specifically, in MRHP task, the matching extent between the review and product description, and inside a review itself (i.e., whether the text and image of a review 
express a similar meaning) impact how customers rate that review---one could probably not contribute kudos unless he finds product-related contents in that review---such contents subsume the confirmation to the seller's  claims, complementary illustration of the product's characteristics, and precautions for the usage, etc.

Based on these two observations and inspired by the idea from relation-based learning~\cite{snell2017prototypical,sung2018learning}, we devise a simple yet effective model, \modelname~(\modelfullname) for MRHP tasks.
~\modelname~gets rid of classic fusion-based architecture and 
use the matching scores between different modalities and fields in various scales as the feature vectors for regression.
Meanwhile, we harness the theory of contrastive learning~\cite{oord2018representation,he2020momentum,chen2020simple} to further boost the model's performance as it has similar mathematical interpretations of relation-based learning. 
To our best knowledge, this is the first work that dedicates to utilizing semantic matching scores as logits for classification.
The contributions of our work are summarized as follows:
\begin{itemize}
    \item We propose~\modelname, a model purely based on semantic matching scores of multi-scale features for the multimodal review helpfulness prediction task.~\modelname~can produce multi-modal multi-field and multi-scale matching scores as expressive features for MRHP tasks.
    \item We design a new functional architecture named aggregation layer, which receives features from a smaller scale and outputs combinations of features in the same scale, plus the counterparts in a larger scale.
    \item We conduct comprehensive experiments on several benchmarks. The results compared with several strong baselines show the great advantage and efficiency of exploiting semantic matching scores for the MRHP task. 
\end{itemize}
\section{Related Work}
\paragraph{Multimodal Representation Learning}
The fundamental solutions of current multimodal tasks focus on multimodal representation learning, which dedicates to extracting and integrating task-related information from the input signals of many modalities~\cite{atrey2010multimodal,ngiam2011multimodal}. 
Recently, multimodal fusion technique becomes the predominant method for expressive representation learning, which coalesces a set of multimodal inputs by 
mathematical operations (e.g., attention and Cartesian product)~\cite{liu2018efficient,tsai2019multimodal,mohla2020fusatnet,hazarika2020misa,han2021bi}.
Though showing exceptional performance on 
those tasks, stacked attention architecture also consumes huge computational power and slows down the training and inference speed.
To alleviate this issue, we devise a fusion-free model for the MRHP task, which escapes from the conventional fusion-based routine.

\paragraph{Relation-based Learning}
The idea of relation-based learning was firstly applied in the few-shot image classification task~\cite{vanschoren2018meta,hospedales2020meta}.
\citet{vinyals2016matching}~employs quantified similarity values between the unseen test images and seen trained images to perform classification.
Prototypical networks~\cite{snell2017prototypical} and Relation Network~\citet{sung2018learning} further treats the correlation matrices between images and pre-computed prototypical feature vectors as logits and optimize them to improve the model's performance.
\citet{lifchitz2019dense}~substitutes the comparison target with implanting weights for better generalization ability.
Later achievements stemming from this theory encompass building up network structures for interactions between samples~\cite{garcia2017few,kim2019edge}, incorporating small-scale computation units like image pixels~\cite{chang2018pyramid,si2018dual,hou2019cross,min2021hypercorrelation} and adding correlation matrices as regularization terms~\cite{wertheimer2021few}.
In the multimodal scenario, semantic matching has also been chosen as the core task to pretrain large multimodal models~\cite{chen2020uniter,kim2021vilt,radford2021learning}.
We inherit this idea to develop a matching-based approach for the MRHP task. 
In our network, sorted matching scores between vectors of different scales and modalities are shaped into regression features. 
We will show this canonical formulation beats many fusion-based strong baselines.

\section{Method}
In this section, we first illustrate the problem definition of \taskfullname~(\taskabbrname). 
Then we elaborate on the model architecture and training process.

\begin{figure*}[ht!]
    \centering
    \includegraphics[trim=0.5cm 0.5cm 2cm 2cm, width=0.92\textwidth]{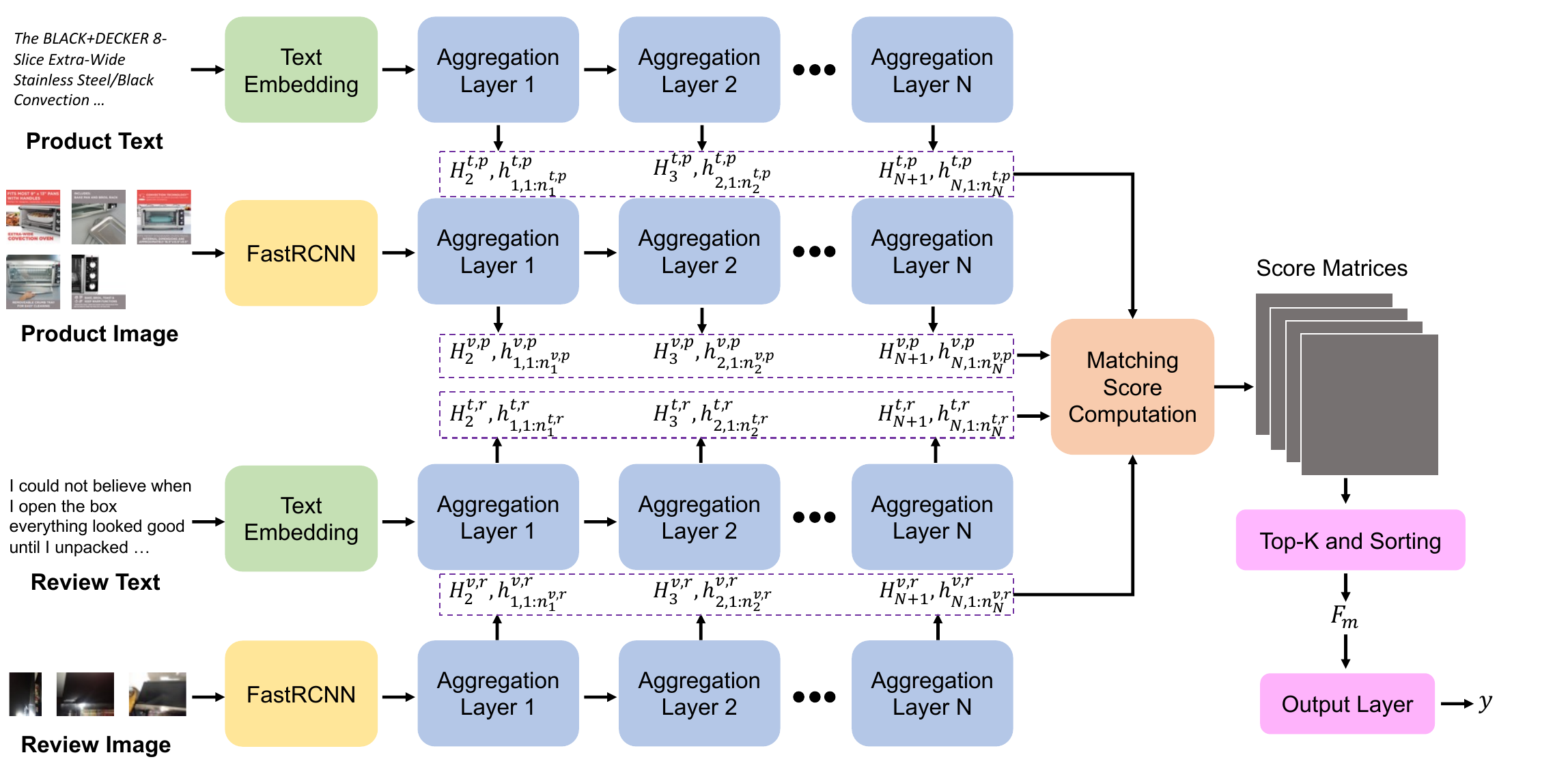}
    \caption{The overall architecture of~\modelname. We hide the data frame reorganization process betwixt two aggregation layers that merge the produced larger-scale representations into a new sequence and only show the outputs from a single block of data with the subscript $i$ omitted for simplicity.}
    \label{model}
\end{figure*}

\subsection{Problem Definition}
Given $N$ product descriptions $\mathcal{P}=\{P_1,P_2,...,P_N\}$ and their associated review sets $\mathcal{R}=\{R_1,R_2,...,R_N\}$, where the review set $R_i$ contains $m_i$ pieces of review $R_i=\{r_{i,1},r_{i,2},...,r_{i,m_i}\}$. 
Both the product descriptions and review pieces
are presented in the modality of text $T_{p_i/r_{i,k}}$ and image $I_{p_i/r_{i,k}}$. 
MRHP aims to predict the helpfulness scores of reviews $\{y_{i,k}\}_{k=1}^{m_i}$ and rank these reviews according to the scores in descending order so that favorable reviews can be promoted to the top.
For the simplicity of the statement, we call the product description and review \textit{field}, denoted by the superscripts $f\in\{p,r\}$. 
Similarly the superscripts $m\in\{t,v\}$ refer to the \textit{modality} of text and image (vision). 

\subsection{Overview}
We depict the overall architecture of our model in~\Cref{model}. 
At the bottom layer, the modality-specific encoders are pretrained models or word vectors that map the raw inputs into continuous embeddings.
The initially embedded representations are viewed as the minimal scale to be aggregated by~\modelname. 
For example, they are word vectors if the encoder is Glove~\cite{pennington2014glove}, contextualized word representations if applying BERT~\cite{devlin2018bert} or other pretrained language models, and detected hot regions in an image when adopting FastRCNN~\cite{Girshick_2015_ICCV}.
These representations are then passed through $N$ stacked aggregation layers where representations from a smaller scale are collated into larger-scale counterparts.
Finally,~\modelname~computes the matching scores between these multi-scale feature vectors and performs regression with the sorted top-K scores.

\subsection{Input Feature}
\paragraph{Textual Representation}
We initialize the token representations of text in the product and review fields with word vectors or pretrained models as
$\mathbf{E'}_t = \{e_1^{'t}, e_2^{'t}, ..., e_l^{'t}\} $, where $l$ is the length of a review sentence. 
For word vector embeddings, we additionally exploit a Gated Recurrent Unit (GRU) \citep{cho2014learning} layer on each sentence to obtain the context-aware token-level representations $\mathbf{E}_t= \{e_1^t, e_2^t, ..., e_l^t\}$,
where $\theta_t$ denotes the parameters of the GRU.

\paragraph{Visual Representation} 
We embed images with pretrained Faster R-CNN \citep{ren2015faster}, which utilizes ResNet-101 as its backbone, yieding the visual feature input $\mathbf{E}_v=\{e_1^v,e_2^v,...,e_{n_h}^v\}$.
where $\theta_v$ denotes the parameters in FastRCNN and $n_h$ is the number of hot regions detected in the given image.


\subsection{Multi-Scale Matching Network (MSMN)}
\begin{figure}[ht!]
    \centering
    \includegraphics[trim=0.7cm 1.7cm 0.9cm 1.4cm, scale=0.3]{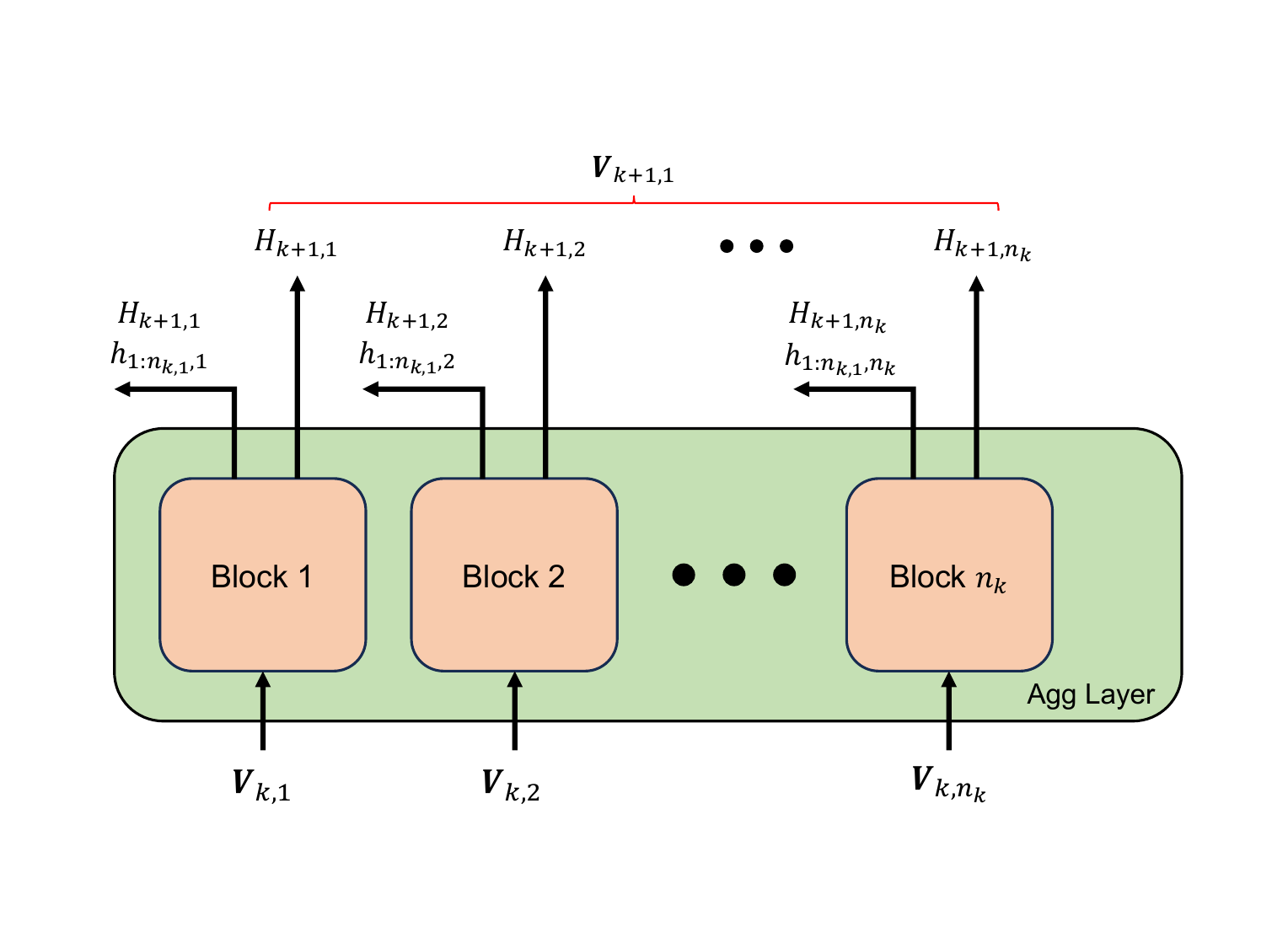}
    \caption{The inner structure of an aggregation layer.}
    \label{layer}
\end{figure}

The inspiration beneath MSMN is from the pyramid and network-in-network architectures~\cite{lazebnik2006beyond,han2021transformer} and relation-based learning~\citep{snell2017prototypical, sung2018learning}.
\paragraph{Multi-scale Feature Generation}
MSMN consists of several structurally-identical aggregation layers that upscale the input sample hierarchically.
An aggregation layer can be further divided into many aggregation blocks, as pictured in~\Cref{layer}, each of which receives the outputs produced from the last layer to generate both the combined representations at the $k$-th scale $h_{1:n_k}=\{h_1,h_2,\hdots,h_{n_k}\}$ of length $n_k$ (for better readablitiy we omit the superscript flags of modality and field) and an aggregated representation at the $k+1$-th (next larger) scale $H_{K+1}$: 
\begin{align}
    \mathbf{V}_{k,i} &= \{H_{k,1},\hdots,H_{k,n_k}\} \\
    H_{k+1,i}, &h_{1:n_k,i} = \mathbf{Aggr}_i(\mathbf{V}_{k,i};\theta_i)
\end{align}
where $\mathbf{V}_{k,i}$ is the collection of the aggregated representations from the $k$-th layer and the subscript $i$ indexes the aggregation block that processes the sequence $i$ of an input instance in the $k$-th layer.
The output representations are all collected to calculate matching scores later, and the upscaled representations are meanwhile gathered as the input sequences for the next layer.
Following~\citet{han2021transformer}, we enforce these internal blocks to share parameters, i.e., $\theta_1=\theta_2=\hdots=\theta_{n_k}$.
In our formulation, we set $N=2$ to endow these scales with realistic meanings (from $k=0$ to $k=2$)---``word $\to$ sentence $\to$ the entire review/description'' for text and ``hot region$\to$ image $\to$ the entire review/description'' for images.

We adopt Transformer~\cite{vaswani2017attention}~as the basic architecture for aggregation layers.
For each layer, We feed the sequential representations from the last layer (after adding the \texttt{[CLS]} token to their heads) into the current layer and extract the heads of the output as the next-scale representations that serve as the input to the next layer:
\begin{align}
    [H_{k+1,i},h_{1:n_k,i}] = \mathbf{Transformer}(\mathbf{V}_{k,i};\mathbf{\Theta})
\end{align}
where $\Theta$ denotes the transformer parameters.

\paragraph{Semantics Refinement}
In the lower layer where the feature scale is small and dense, there are closed semantic units, which result in many duplicated matching scores and impair the ability of prediction network (as we show in the next section)
To address this problem, we aim to filter the extremely long sequences of output features, only to maintain the dominant components.
The algorithm is based on a faster $k$-means algorithm, which can produce a set of representations by clustering adjacent points so that redundant semantics are eliminated.
To reduce extra overhead, we implement an approximate but faster algorithm~\cite{hamerly2010making}~only when the feature number exceeds a threshold.
To prevent each clustered set from being too small (i.e., only 1 or 2 points) and ensure efficiency, we random sample $C$ points as centers.
The algorithm is formally depicted in~\Cref{alg:1}, where we omit the specific steps of fast $k$-means and readers can refer to the paper~\cite{hamerly2010making} for the details. It should be emphasized that $k$-means is a non-parametric clustering algorithm and only incurs negligible overhead in the forward pass.
\begin{algorithm}[h!]
    \small
    \SetKwFunction{ret}{return}
    \SetKwFunction{RS}{RandomSample}
    \SetKwFunction{RI}{RandomInit}
    \SetKwFunction{KM}{$k$-Means}
    \SetAlgoLined
    \KwIn{Semantic elements set $S=\{s_1,s_2,...,s_N\}$, expected cluster size $r$, number of centers $C$, }
    \KwOut{Refined set $S'=\{s'_1,s'_2,...,s'_{C}\}$}
    \BlankLine
    \uIf{$N \leq C \times r$}
    {\ret \RS$(S,C)$ \\
    }\Else{
        \ret $\KM(\RI(C),S)$ \\
    }
\caption{\textbf{Semantics Refinement}}
\label{alg:1}
\end{algorithm}

In the algorithm, $C$ bounds the lowest number of centers required for the later computation and a typical value could be $C=\lceil \sqrt{K} \rceil$ where $K$ is the hyperparameter in the last layer's feature selector (as stated below) implicitly and roughly. The heuristics in using squareroot comes from the fact that similarity scores are computed in pair. 
Here $r$ is another important hyperparameter that controls the expected (or average) cluster size to avoid overmuch small clusters.

\paragraph{Prediction} After obtaining the representations from both fields and modalities in all scales ($H_k, h_{1:n_k}$), we concatenate them into four matrices $\mathbf{R}^{t,p},\mathbf{R}^{t,r},\mathbf{R}^{v,p},\mathbf{R}^{v, r}$ whose rows are these representation vectors.
Concretely, we extract the n-gram token, sentence, and n-gram sentence representations for text, and n-gram RoI and image representations for the image.
We hypothesize that the review quality depends on the semantic coherence existing within 
1) The image-text pair in the review to guarantee the coherence of the review itself. Low-quality reviews are usually not self-contained ($\mathbf{R}^{t,r},\mathbf{R}^{v,r}$). 
We exclude the scores of image-text pairs from product introduction because they do not have any impact to the helpfulness of a review.
2) The same modality from different fields ($\mathbf{R}^{t,p},\mathbf{R}^{t,r}$ and $\mathbf{R}^{v,p},\mathbf{R}^{v,r}$). This is important because user-preferred comments should directly response to the selling points in the introduction.

The matching scores are calculated as the cosine similarities between row vectors of two matrices:
\begin{equation}
    \label{eq:cos_sim}
    \textbf{S}(\mathbf{A},\mathbf{B}) = \mathbf{cosine}(\mathbf{A},\mathbf{B}) = \frac{\mathbf{A}\mathbf{B}^T}{\Vert \mathbf{A} \Vert\cdot\Vert \mathbf{B} \Vert^T} \\
\end{equation}
where $\Vert \cdot \Vert$ is the row-wise L2 normalization.
Suppose there are $n_1,n_2,n_3,n_4$ rows (the number of feature vectors) in the four matrices, then we have $n_1n_2+n_2n_4+n_3n_4$ matching scores.
We then picked the highest $K$ scores to form the last features. 
It should be emphasized that the top-K operation reorganizes the tensors during its computation. 
\begin{equation}
    \mathbf{h} = \mathbf{TopK}(\mathbf{FlattenAll}(\mathbf{S'}))
\end{equation}
Therefore the gradient back-propagation path is not constant given different input samples. The predictions for training and inference are calculated from the feature:
\begin{equation}
    \mathbf{f}_{i,j} = \sigma(\mathrm{Linear}(\mathbf{h}_{i,j}))
\end{equation}
where $\sigma$ is the Sigmoid function.

\subsection{Training}
\label{sec:Training}
We follow \citet{nguyen2023gradient}~to apply the listwise loss for training.
\begin{equation}
    \mathcal{L} = -\sum_{i=1}^{\vert P \vert}\sum_{j=1}^{\vert R_i \vert}y_{i,j}'\log(f_{i,j}')
\end{equation}
where $\vert \mathcal{P} \vert$ is the number of productions in the batch and $\vert R_i \vert$ is the number of reviews corresponding to $P_i$. The normalized labels $y'$ and predictions $f'$ are given by
\begin{equation}
    f'_{i,j} = \mathrm{softmax}(\mathbf{f}_i)_j,
    y'_{i,j} = \mathrm{softmax}(\mathbf{y}_i)_j
\end{equation}
Note that the final predictions are ranged within $(0, 1)$, which diverges from the true label distribution $y \in [0,4]$. However, the ultimate target (same as the evaluation metrics) of the task concentrates on \textit{ranking} (relative value) rather than absolute value. This kind of learning can still benefit the performance.

\begin{table}[h]
    \centering
    \small
    \resizebox{\linewidth}{!}{
    \begin{tabular}{c|c|ccc}
    \toprule
         \multirow{3}{*}{Model}
        & \multirow{3}{*}{\#Params} 
        & \multirow{3}{*}{\shortstack{Statistical \\ Learning }} & \multirow{3}{*}{Fusion} & \multirow{3}{*}{\shortstack{Matching \\ Score \\ as features}} \\
         ~ & & & \\
         ~ & & & \\
         \hline
         SSE-Cross  & — & \xmark & \cmark & \xmark \\
         D\&R Net  & — & \cmark & \cmark & \xmark \\
         MCR & 2.33M & \xmark & \cmark & \xmark \\
         SANCL & 2.38M & \cmark & \cmark & \xmark \\
         CMCR & 2.41M & \xmark & \cmark & \xmark \\
         GBDT & 21.8M & \xmark & \cmark & \xmark \\
         \modelname  & 2.28M &  \xmark & \xmark  & \cmark \\
        \bottomrule
    \end{tabular}
    }
    \caption{Comparison between baseline models and ours.}
    \label{tab:model_comp}
\end{table}
\renewcommand{\arraystretch}{1.1}
\begin{table*}[ht]
    \small
    \resizebox{\textwidth}{!}{
    \begin{tabular}{l*{3}{|c c c}}
        \toprule
        \multirow{2}{*}{Model} & 
        \multicolumn{3}{c|}{Cloth. \& Jew.} & \multicolumn{3}{c|}{Electronics}  & \multicolumn{3}{c}{Home \& Kitchen}  \\
         ~ & \textbf{MAP} & \textbf{N@3} & \textbf{N@5} & \textbf{MAP} & \textbf{N@3} & \textbf{N@5} & \textbf{MAP} & \textbf{N@3} & \textbf{N@5} \\
        \midrule
        SSE-Cross & 65.0 & 56.0 & 59.1 & 53.7 & 43.8 & 47.2 & 60.8 & 51.0 & 54.0 \\
        D\&R Net &  65.2 & 56.1 & 59.2 & 53.9 & 44.2 & 47.5 & 61.2 & 51.8 & 54.6 \\
        MCR  & 66.4 & 57.3 & 60.2 & 54.4 & 45.0 & 48.1 & 62.6 & 53.5 & 56.6  \\
        SANCL & 67.3 & 58.6 & 61.5 & 56.2 & 47.0 & 49.9 & 63.4 & 54.3 & 57.4 \\
        CMCR & 67.4 & 58.6 & 61.6 & 56.5 & 47.6 & 50.8 & 63.5 & 54.6 & 57.8 \\
        GBDT & 82.6 & 80.3 & 79.3 & 74.2 & 68.0 & 69.8 & 81.7 & 76.5 & 78.8 \\
        \modelname~(Ours) & \textbf{92.3} & \textbf{90.4} & \textbf{91.5} & \textbf{81.4} & \textbf{78.6} & \textbf{75.6} & \textbf{88.6} & \textbf{88.3} & \textbf{88.4}  \\
        \hline
        MCR+BERT & 65.8 & 55.9 & 58.8 & 55.9 & 46.8 & 49.4 & 62.4 & 52.9 & 56.1 \\
        GBDT+BERT & 80.3 & 78.7 & 77.1 & 73.8 & 68.3 & 69.5 & 81.4 & 76.9 & 79.4\\
        \modelname~+BERT~(Ours) & \textbf{91.5} & \textbf{89.7} & \textbf{90.1} & \textbf{79.3} & \textbf{77.7}& \textbf{78.6} & \textbf{87.7} & \textbf{85.9} & \textbf{86.2} \\
        \bottomrule
    \end{tabular}
    }
    \caption{Results on the Amazon-MRHP (English) dataset. All reported metrics are the average of five runs. Baseline results are from \citet{nguyen2023gradient}. PREMISE outperform the strongest baseline with p-value<0.05 based on the paired t-test.}
    \label{amazon results}
\end{table*}

\renewcommand{\arraystretch}{1.05}
\begin{table*}[ht]
    \centering
    \small
    \resizebox{\linewidth}{!}{
    \begin{tabular}{l*{3}{|c c c}}
        \toprule
         \multirow{2}{*}{Model} &
         \multicolumn{3}{c|}{Cloth. \& Jew} & \multicolumn{3}{c|}{Electronics}  & \multicolumn{3}{c}{Home \& Kitchen}   \\
        ~ & \textbf{MAP} & \textbf{N@3} & \textbf{N@5} & \textbf{MAP} & \textbf{N@3} & \textbf{N@5} & \textbf{MAP} & \textbf{N@3} & \textbf{N@5} \\
        \midrule
        SSE-Cross & 66.1 & 59.7 & 64.8 & 76.0 & 68.9 & 73.8 & 72.2 & 66.0 & 71.0 \\
        D\&R Net &  66.6 & 61.3 & 65.8 & 76.5 & 69.5 & 74.3 & 72.7 & 66.7 & 71.8 \\
        MCR &  68.8 & 62.3 & 67.0 & 76.8 & 70.7 & 75.0 & 73.8 & 67.0 & 72.2  \\
        SANCL & 70.2 & 64.6 & 68.8 & 77.8 & 71.5 & 76.1 & 75.1 & 68.4 & 73.3 \\
        CMCR & 70.3 & 64.7 & 69.0 & 78.2 & 72.4 & 79.6 & 75.2 & 68.8 & 73.7 \\
        GBDT & 78.5 & 77.1 & 79.0 & 87.9 & 86.7 & 88.1 & 85.6 & 78.8 & 83.1 \\
        \modelname~(Ours) &  \textbf{95.3} & \textbf{94.6} & \textbf{95.0}  & \textbf{96.9} & \textbf{96.1} & \textbf{96.8} & \textbf{93.9} & \textbf{91.5} & \textbf{92.8} \\
        \bottomrule
    \end{tabular}
    }
    \caption{Results on the Lazada-MRHP (Indonesian) dataset. 
    }
    \label{lazada results}
\end{table*}

\section{Experiments}
\subsection{Datasets and Metrics}
We evaluate our model on two \taskabbrname~datasets \citep{liu2021multi}, each of which subsumes same three categories: \textit{Clothing, Shoes \& Jewelry, Home \& Kitchen} and \textit{Electronics}. We train and test both datasets on a single NVIDIA RTX A6000 GPU.
The gradients are calculated and backpropagated for each batch in a single forward pass, without batch division and gradient accumulation.
We compare our model with several baseline models on three common metrics for ranking tasks: the mean average precision (MAP), the N-term ($N=3,5$ in our experiment in accord with previous works) Normalized Discounted Cumulative Gain (NDCG@N)~\citep{jarvelin2017ir, diaz2018modeling}.
The helpfulness scores are labeled as the logarithm of approval votes to the corresponding reviews and are clipped into integers within $[0,4]$. 
The statistics of datasets and more training details are provided in appendix.

\subsection{Baselines}
We compare our model with the following baselines: Stochastic Shared Embeddings enhanced cross-modal network (SSE-Cross)~\cite{abavisani2020multimodal}, Decomposition and Relation Network~(D\&R Net)~\cite{xu2020reasoning}. The Multimodal Coherence Reasoning network~(MCR)~\cite{liu2021multi} designs several reasoning modules based on fused representations for prediction. SANCL~\cite{han-etal-2022-sancl} and contrastive-MCR (CMCR)~\cite{nguyen-etal-2022-adaptive} minimize auxiliary contrastive loss to refine the multimodal representations. 
Gradient-boosted decision tree (GBDT)~\cite{nguyen2023gradient} design a random walk policy and aggregate the helpfulness scores through from tree leaves---the endpoints of the random walk.

To provide a holistic view on the distinctions between the learning paradigms of these models, we list and compare three key characteristics between~\modelname~and baselines in Table~\ref{tab:model_comp}.
From the table we observe that all baseline models contain fusion modules inside the entire structures.
Moreover, D\&R Net and SANCL also incorporate extra statistical correlations (Adjective-Noun Pairs and selective-attention mask creation) that inject external knowledge to bridge the semantic gap between textual and visual modality or cast more focus on contents that perceived important by human beings.
Our model escapes from both complicated manually crafted features and conventional model architecture by directly computing the matching scores and automatically picking the $K$ highest ones as features for regression. 

\subsection{Results}
We run our models three times and report the average performance in Table~\ref{amazon results}~and~\ref{lazada results}.
It can be clearly seen that our model outperforms all these baselines on two datasets.
Particularly, compared with the strongest baseline---GBDT~\cite{nguyen2023gradient}, PREMISE gains over 5 points improvement on MAP and NDCG@5 and 10 points improvement on NDCG@3 on Amazon-MRHP dataset, and 6.8$\sim$17.5 improvement on all metrics on Lazada-MRHP datasets.
When using BERT to initialize embeddings, we note a slight performance degradation compared to the implementations that use GloVe as embeddings in both~\modelname and other baselines. 
Such outcome demonstrates the superiority of our fusion-free model and, at least in the MRHP task, multimodal fusion is not a necessity and may hinder the model from better performance.

Besides, we highlight the size of the feature vectors in the last layer.
Fusion-based baselines usually concatenate representations from both fields and modalities to perform final regression, which requires at least 512 (128$\times$4) dimensions of feature vectors.
The vector is even longer in MCR and SANCL (over 1000) since there are many extra features taken into account.
Nevertheless, the feature vector lengths for the best performance in~\modelname~are apparently smaller.
As shown in~\Cref{fig:perf_kvalue}, the optimal choices of $K$ range from 64 to 128.
This fact shows that there could be many redundant elements in the vectors generated by fusion-based models, and~\modelname~successfully enhance the efficiency of unit length features through a simple representation learning policy.

\begin{table}[ht]
    \small
    \centering
    \resizebox{\linewidth}{!}{
    \begin{tabular}{l|*{3}{c}}
        \toprule
        Description & \textbf{MAP} & \textbf{N@3} & \textbf{N@5} 
        \\
        \midrule
        \modelname~(Amazon) &  \textbf{87.4} & \textbf{85.8} & \textbf{86.5} 
        \\
        \midrule
            -w/o n-gram token repr & 84.8 & 82.1 & 83.3 \\
            -w/o sent repr & 86.2 & 84.3 & 85.0  \\
            -w/o n-gram sent repr & 85.7 & 83.9 & 84.6 \\
            -w/o n-gram RoI repr & 83.9 & 81.8 & 82.6 \\
            -w/o image repr & 86.5 & 84.1 & 85.3 \\
            -w/o n-gram token \& n-gram RoI repr & 75.3 & 69.8 & 72.2 \\
            -w/o n-gram sent repr \& image repr & 84.1 & 82.5 & 83.0 \\
        \midrule
        \modelname~(Lazada) &  \textbf{95.4} & \textbf{94.0} & \textbf{94.9} 
        \\
        \midrule
            -w/o n-gram token repr & 91.0 & 88.9 & 89.5 \\
            -w/o sent repr & 93.5 & 91.6 & 92.2 \\
            -w/o n-gram sent repr & 94.3 & 92.9 & 93.8 \\
            -w/o n-gram RoI repr & 92.7 & 90.1 & 91.7 \\
            -w/o image repr & 94.8 & 94.2 & 94.6 \\
            -w/o n-gram token \& n-gram RoI repr & 80.1 & 78.6 & 79.5  \\
            -w/o n-gram sent repr \& image repr & 92.3 & 89.6 & 90.1 \\
        \bottomrule
    \end{tabular}
    }
    \caption{Ablation experiments of~\modelname~on two datasets. The values are averaged over all three categories.}
    \label{tab:abl}
\end{table}

\subsection{Ablation Study}
We run our models several ablative settings under feature selection.
These settings correspond to excluding some features from different scales when computing the multiscale matching scores;
During implementation, we mask out these vectors. 

The results are summarized in Table~\ref{tab:abl}, from which we have the following discoveries.
First, discarding representations of any scale causes degradation in the model's performance, indicating that all these chosen features contribute to the accurate prediction.
In addition, performance plummets more severely when features are removed on smaller scales (or in bottom layers), including a single scale (e.g. ``n-gram token repr" vs. ``n-gram sent repr", ``n-gram RoI repr" v.s. ``n-gram image repr") or the combinations (e.g., ``n-gram token \& n-gram RoI repr" v.s. ``n-gram sent repr \& image repr``). 
This outcome reveals that lower-level feature is more fundamental to the model's performance than higher ones, since a large portion of matching scores are computed from them.

\section{Analysis}
\subsection{The Impact of Selected Feature Numbers}
\begin{figure}[ht!]
    \centering
    \includegraphics[scale=0.5]{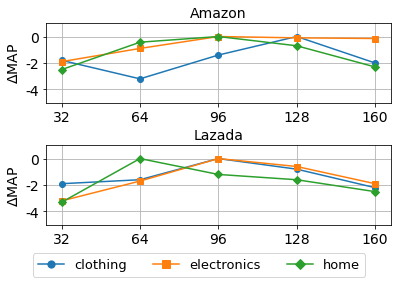}
    \caption{The relative MAP drop (the absolute value of $\Delta$MAP) from the optim to different $K$.
    Performance when $K>160$ or $K<32$ is far lower than the optimum so we do not include in the figure.}
    \label{fig:perf_kvalue}
\end{figure}

A unique hyperparameter in~\modelname~could be $K$ for the selection of last features which determines how many highest scores to be included to form the final feature vectors.
To explore how $K$ affects the model's performance, we run our model with various values of $K$ and plot how the performance changes in Figure~\ref{fig:perf_kvalue}.
It can be found that to achieve ideal performance in both datasets, an appropriate choice of $K$ (from 64 to 128) is necessary. 
When setting $K$ too high or too low, i.e., dispersing the model's concentration on too many scores or forcing the model to focus on only a few highest scores, the model fails to reach the optimum. 
This phenomenon reveals that a promising filter should provide comprehensive coverage of the matching scores for the prediction layer.

\subsection{The Impact of Lower Layer Filter}
\begin{figure}
    \centering
    \includegraphics[scale=0.5]{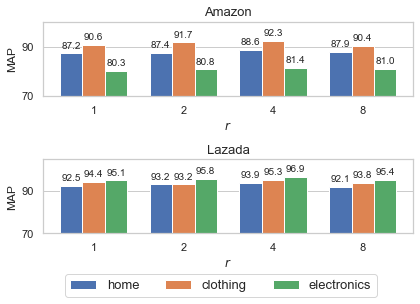}
    \caption{The performance (MAP) under different $r$ on two datasets.}
    \label{fig:perf_cluster}
\end{figure}

Apart from the last-layer feature selector, we also insert many filters into the lower layers. 
To verify the efficacy of this design, we performed additional experiments by varying $r_{min}$~in~\cref{alg:1} while fixing $K=96$ and $C=\lceil\sqrt{K}\rceil=10$. 
The results on the two datasets are shown in~\Cref{fig:perf_cluster}, from which we notice that in all categories, a proper choice of $r$ value ($k=4$ in our experiments) can further enhance the performance by removing duplicated semantics in lower aggregation layers. 
This suggests that the semantics redundancy removal procedure the combination of $k$-means and random sampling can serve as a primary filter for the feature selection.

\subsection{Why Does BERT Fail?}
\label{sec:bert_fail}
As mentioned above, it is weird that after replacing the word vectors (GloVe and FastText) with the pretrained language model in the embedding layer, both the fusion-based and fusion-free models fail to produce a significant increase as in other multimodal tasks. 
We surmise that this is mainly due to informal text input.
Upon manual inspection, we find many pieces of low-quality reviews---especially those of low helpfulness scores.
Take review 2 in Table~\ref{Problem Setting} as an example, 
the review passage is readable by sentence except for some grammatical errors, but the logic is messy and out of the topic. 
The results of previous work on tasks related to spoken language (informal text) have shown that BERT may not lead to a performance improvement~\citep{gu2020data}.
To further verify our hypothesis, we carry out a group of blank control experiments on Amazon-MRHP dataset.
Specifically, we run regression directly on: A) representations encoded by a single layer GRU with Glove 300d as word embeddings in both fields; B) the representations at the position of [CLS] token using BERT-base-uncased as the pretrained encoder. 
The results are shown in Table~\ref{tab:bert_studyl}.
From the table we find the performance between word vectors and BERT pretrained models are very close. 
This outcome looks consistent with the results in~\citet{gu2020data} and may substantiate our aforementioned hypothesis.

\begin{table}[h]
    \centering
    \begin{tabular}{c|c|c|c|c}
        \toprule
        Category & Setting & MAP & N@3 & N@5 \\
        \midrule
        \multirow{2}{*}{Clothing} & A & 64.83 & 55.62 & 58.95   \\
         ~ & B & 64.75 & 55.51 & 59.03   \\
        \midrule 
        \multirow{2}{*}{Electronics} & A & 53.63 & 43.77 & 47.31  \\
         ~ & B & 53.90 & 43.85 & 47.02 \\
         \midrule
        \multirow{2}{*}{Home} & A & 61.08 & 51.17 & 54.26 \\
         ~ & B & 61.03 & 51.09 & 54.14 \\
         \bottomrule
    \end{tabular}
    \caption{A group of blank control experiments on Amazon-MRHP dataset.}
    \label{tab:bert_studyl}
\end{table}

\section{Conclusion}
In this work, we propose a novel matching-based learning  model, \modelname, for the task of multimodal review helpfulness prediction (MRHP).
\modelname~calculates matching scores between refined semantics across modalities and data fields for fast and accurate regression and ranking.
Experiments and analysis demonstrate that our model exceeds many strong fusion-based approaches, which provides a possible idea for such kind of tasks.
\section*{Limitations}
The major limitation of~\modelname~is its applicable scenarios or restricted adaptation ability to other multimodal tasks.
Ideally, we expect~\modelname~to behave as a generic model that can also work on many other multimodal tasks, but now we have only empirically demonstrated its efficacy in the MRHP task. 
Intuitively, we believe that at least in the tasks where the extent of semantic matching matters, our method should produce satisfying results, e.g., multimodal (image/text) retrieval and sarcasm detection where low correlation usually implies that sarcasm exists.
But currently we only yield fair results that fall behind the current SOTA significantly on the aforementioned tasks (see appendix for details).

Another limitation is the efficiency. We actually adopt a brute-force computing strategy, which can be further improved through more careful module design. We hold this as our future potential direction to work on.

\bibliography{anthology,custom}
\bibliographystyle{acl_natbib}

\appendix

\appendix

\section{Dataset Specifications}
We list the specifications of train/validation/test split of the two datasets (six categories) in Table~\ref{Amazon-Spec}~and Table~\ref{Lazada-Spec}.
The numbers ``$X/Y$'' represent that the split contains $X$ product descriptions and $Y$ reviews.
Amazon-MRHP is an pure English dataset, while Lazada-MRHP is written in Indonesian so that we do not conduct BERT related experiments on it.
\begin{table}[ht]
    \centering
    \small
    \resizebox{\linewidth}{!}{
    \begin{tabularx}{0.5\textwidth}{lccc}
        \toprule
        \multicolumn{4}{c}{\textbf{Amazon-MRHP (Products/Reviews)}} \\
        [0.5ex]
        Cat. & Cloth. \& Jew. & Elec. & Home \& Kitch.\\
        \midrule
        Train & 12,074/277,308  & 10,564/240,505 & 14,570/369,518 \\
        Dev &  3,019/122,148 & 2,641/84,402 & 3,616/92,707 \\
        Test & 3,966/87,492 & 3,327/79,750  &  4,529/111,593 \\
        \bottomrule
    \end{tabularx}
    }
    \caption{Statistics of the Amazon-MRHP dataset.}
    \label{Amazon-Spec}
\end{table}

\begin{table}[ht]
    \centering
    \small
    \resizebox{\linewidth}{!}{
    \begin{tabularx}{0.5\textwidth}{lccc}
        \toprule
        \multicolumn{4}{c}{\textbf{Lazada-MRHP (Products/Reviews)}} \\
        [0.5ex]
        Cat. & Cloth. \& Jew. & Elec. & Home \& Kitch.\\
        \midrule
        Train & 6,596/10,4093 & 3,848/41,828 &  2,939/36,991 \\
        Dev & 1,649/26,139 & 963/10,565 & 736/9,611 \\
        Test & 2,062/32,274 & 1,204/12,661 & 920/12,551 \\
        \bottomrule
    \end{tabularx}
    }
    \caption{Statistics of the Lazada-MRHP dataset.}
    \label{Lazada-Spec}
\end{table}

\section{Training Details}
\subsection{Initialization of Embeddings}
To stay consistent with previous works, we embed the text input of Amazon-MRHP with GloVe-300d~\cite{pennington2014glove} and Lazada-MRHP with  Fasttext~\cite{joulin2016fasttext}, respectively.
In BERT-related experiments we employ the Huggingface toolkit for pretrained models\footnote{https://huggingface.co/docs/transformers/index}.

\subsection{Hyperparameter Search space}
The optimal hyperparameter settings are listed in Table \ref{Amazon hyperparameters}, \ref{lazada hyperparameters} and \ref{Amazon Bert hyperparameters}.
The search space of these hyperparameters are: learning rate in $\{1e^{-4}, 5e^{-4}\}$, text embedding dropout fixed at $\{0.2\}$, shared space hidden dimension in $\{128, 256\}$.

\begin{table}[ht!]
    \centering
    \resizebox{\linewidth}{!}{\begin{tabular}{lccc}
    \toprule
    \multicolumn{4}{c}{\textbf{Amazon-MRHP Glove Hyperparameters}} \\
         & Cloth. \& Jew. &  Elec. & Home \& Kitch. \\
         \midrule
         learning rate  & $1e^{-4}$  & $5e^{-4}$ & $5e^{-4}$ \\
         text embedding dim & 300 & 300 & 300 \\
         text embedding dropout & 0.2 & 0.2 & 0.2 \\
         image embedding dim & 256 & 256 & 256 \\
         text embedding dim & 128 & 128 & 128 \\
         shared space hidden & 128 & 128 & 128 \\
         $r$ & 4 & 4 & 4 \\
         K & 96 & 128 & 64 \\
         batch size & 32 & 32 & 32 \\
         \bottomrule
    \end{tabular}
    }
    \caption{Hyperparameters for Amazon-MRHP using glove-300d embeddings.}
    \label{Amazon hyperparameters}
\end{table}

\begin{table}[ht!]
    \centering
    \resizebox{\linewidth}{!}{\begin{tabular}{lccc}
    \toprule
    \multicolumn{4}{c}{\textbf{Lazada-MRHP fastText Hyperparameters}} \\
         & Cloth. \& Jew. &  Elec. & Home \& Kitch. \\
         \midrule
         learning rate  & $1e^{-4}$  & $5e^{-4}$ & $1e^{-4}$ \\
         text embedding dropout & 0.2 & 0.2 & 0.2 \\
         image embedding dim & 256 & 256 & 256 \\
         text embedding dim & 128 & 128 & 128 \\
         shared space hidden & 128 & 128 & 128 \\
          $r$ & 4 & 4 & 4 \\
          $K$ & 96 & 96 & 128 \\
         batch size & 32 & 32 & 32 \\
         \bottomrule
    \end{tabular}
    }
    \caption{Hyperparameters for Lazada-MRHP using fasttext embeddings.}
    \label{lazada hyperparameters}
\end{table}

\begin{table}[ht!]
    \centering
    \resizebox{\linewidth}{!}{\begin{tabular}{lccc}
    \toprule
    \multicolumn{4}{c}{\textbf{Amazon-MRHP BERT Hyperparameters}} \\
         & Cloth. \& Jew. &  Elec. & Home \& Kitch. \\
         \midrule
         learning rate  & $1e^{-5}$  & $1e^{-5}$ & $1e^{-5}$ \\
         bert learning rate &  $1e^{-5}$  & $1e^{-5}$ & $1e^{-5}$ \\
         text embedding dropout & 0.2 & 0.2 & 0.2 \\
         image embedding dim & 512 & 512 & 512 \\
         text embedding dim & 256 & 256 & 256 \\
         shared space hidden & 256 & 256 & 256 \\
          $r$ & 4 & 4 & 4 \\
          $K$ & 128 & 128 & 128 \\
         batch size & 16 & 16 & 8 \\
         \bottomrule
    \end{tabular}
    }
    \caption{Hyperparameters for all categories using BERT as encoder}
    \label{Amazon Bert hyperparameters}
\end{table}

\label{sec:appendix}

\subsection{Sampling of Production Description-Review Pairs in Training}
We mentioned in~\cref{sec:Training} that the training pairs are sampled from the training set.
Now we describe how do we sample these training pairs.
First, we sample $B$ products from the training set where $B$ is the batch size.
Next, for each product, we randomly sample one of its positive review (rating is greater than 2 and $N_r^{-}$ negative reviews (rating is less than or equal to 2 from the corresponding review set.
The dataset has been filtered during manufacture time so that there is always at least one positive/negative review under each product.
To put it in a nutshell, a sampled batch contains $B$ product descriptions, $B$ positive reviews and $N_r^{-}B$ negative reviews.

\section{The Differentiability of top-K Operation in PyTorch}
Given a vector $S = \{s_1, s_2, ..., s_L\} \in \mathbb{R}^L$ where $L$ is the length of that vector, when passing through the top-K operation, most fundamentally its largest $K$ values are selected and sorted in descending order to form a new vector $T = \{t_1, t_2, ..., t_K\} \in \mathbb{R}^K$. 
Suppose the indices of $T$'s elements in $S$ are $I= \{i_1, i_2, ..., i_K\} \subset \{1, 2, ..., L\}$, then the process equals to that concurrently there is a mask $M \in \{0,1\}^L $ automatically created and ``multiplies'' on $S$. 
Each value $m_j$ in $M$ is
\begin{equation*}
    m = 
    \begin{cases}
        1, \quad if \quad j \in I \\
        0, \quad if \quad j \notin I
    \end{cases}
\end{equation*}
With this mask, the forward and backward propagation can proceed as in conventional routines.

\section{Training Speed}
\subsection{Theoretical Analysis}
For simplicity, we consider the case of a pair of modality sequence.
Let $X_1\in \mathbb{R}^{l_1\times d}$, $X_2\in \mathbb{R}^{l_2\times d}$ be a pair of input modality sequences. Here we assume they have been both projected to the same dimension as a common practice that both the fusion--prediction routines and our matching approach exercise.
The multihead attention operation in fusion-based models can be written as:
\begin{equation}
    X_{12} = Att(X_1,X_2)
\end{equation}
Note that attention is a \textit{directional} operation, i.e. $Att(M_1,M_2)\neq Att(M_2,M_1)$. 
Due to this, a fusion-based learning model $\mathcal{M}$ always adopts a pair of conjugate attention. Therefore, for a model of $N$ layers, the total computational complexity $C_f$ is:
\begin{equation}
    C_{f}=(2l_1l_2+l_1^2+l_2^2)Nd
\end{equation}

Now consider matching-based models. 
We only have self-attention for each modality per layer. 
The whole computational complexity consists of the self-attention (att) and multi-scale matching score (mm).
\begin{equation}
    C_{m} = C_{att} + C_{mm}
\end{equation}
Since the number of scales decreases as the aggregation proceeds, we denote the decreasing ratio at layer $i$ for modality $j$ as $k_{i,j}$. Hence, the total computational complexity is:
\begin{equation}
    C_{att}=2\sum_pl_p^2d\left(1+\frac{1}{k_{p,1}^2}+\frac{1}{k_{1,p}^2k_{2,p}^2}+...\right) 
\end{equation}
In our settings, $k_{p,1}$ is large (typically greater than 10), therefore $\frac{1}{k_{p,j}^2} < 0.01$ and can be ignored:
\begin{equation}
    C_{att} = 2(l_1^2+l_2^2)d
\end{equation}
As for the second term, we have:
\begin{equation}
\begin{split}
    C_{mm} &=  l_1\left(\frac{1}{k_{1,1}}+...\right)l_2\left(\frac{1}{k_{2,1}}+...\right)d \\
    &< \frac{l_1l_2d}{k_{1,1}k_{2,1}(1-k_1^{-1})(1-k_2^{-1})} \\
\end{split}
\end{equation}
In the MRHP dataset, $l_1\approx l_2 = l$. 
For the typical value $N=2, k_1=k_2=10$, we have $C_f=8l^2d$ and $C_m < (4+\frac{1}{81})l^2d=4.01l^2d$, or 
\begin{equation}
    \frac{C_m}{C_f}\approx0.5
\end{equation}
which is closed to the measured acceleration in~\ref{fig:speedcomp}. In fact, let $C_m=C_f$ and $N=2$, we have $l_1\approx 2.42 l_2$, which seldom happens in the whole dataset. We observe that the number of hot regions is greater than the text length.

\subsection{Numerical Results}
\begin{figure}
    \centering
    \resizebox{\linewidth}{!}{
        \includegraphics{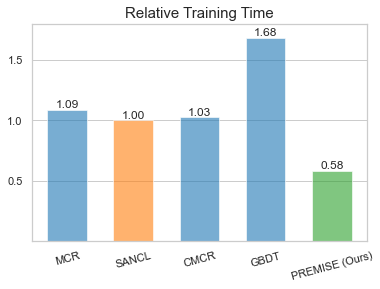}
    }
    \caption{The relative training time of different models. The fastest baseline (SANCL) is highlighted in orange, while our model is highlighted in green. Others are in blue.}
    \label{fig:speedcomp}
\end{figure}

We measure the average training time per batch of MCR, SANCL (the state-of-the-art baseline) and PREMISE, as shown in Figure~\ref{fig:speedcomp}.
The average values are calculated by counting the total time of iterations over 100 batches for 5 random intervals during the whole training process. 

Mathematically, denote the counted time of the $i^{th}$ interval as $t_i$, the speed is calculated as follows
\begin{equation}
    speed = \frac{\sum_{i=1}^5 t_i}{100 \times 5} 
\end{equation}
It can be seen that the training time has been greatly shortened by 42\% and 65\% compared to SANCL (the fastest baseline, athough they have closed number of parameters as shown in~\Cref{tab:model_comp}) and GBDT (the strongest baseline), which approximately matches the conclusion given by mathematical deduction. 

\section{Experiment on Multimodal Retrieval}
We test~\modelname~on multimodal retrieval (bidirectional) task, the results of both image-to-text and text-to-image retrieval on the MSCOCO test set are shown in~\Cref{tab:retrieval}. It can be seen that although our formulation process is completely based on ``learning-from-relation" in MRHP task, the constructed model still has some generalizability to other tasks that our hypothesis stands.

\begin{table*}[t]
    \centering
    \begin{tabular}{c|c|c|c|c}
    \toprule
       \multirow{2}{*}{Model}  & \multicolumn{2}{c|}{Image-to-Text} & \multicolumn{2}{c}{Text-to-Image} \\
    \cline{2-5}
    ~ & PRMP & R@1 & PRMP & R@1 \\ 
    \hline
    VCRN~\cite{li2019visual} & 29.70 & \textbf{53.00} & 29.90 & \textbf{40.50} \\
    PCME~\cite{chun2021probabilistic} & 34.10 & 41.70 & \underline{34.40} & 31.20 \\
    MSRM~\cite{wang2023multilateral} & \textbf{35.62} & \underline{44.32} & \textbf{35.81} & \underline{33.40} \\
    \hline
    \modelname & \underline{34.23} & 42.06 & 33.92 & 31.50 \\
    \bottomrule
    \end{tabular}
    \caption{Results on MSCOCO-5K test set. The highest values in each metric are in bold, while the second-highest are indicated with an underline.}
    \label{tab:retrieval}
\end{table*}

\section{Case Study}
\subsection{How matching scores affect the prediction?}
\begin{figure*}
    \centering
    \includegraphics[scale=0.52, trim=2cm 0 0 0]{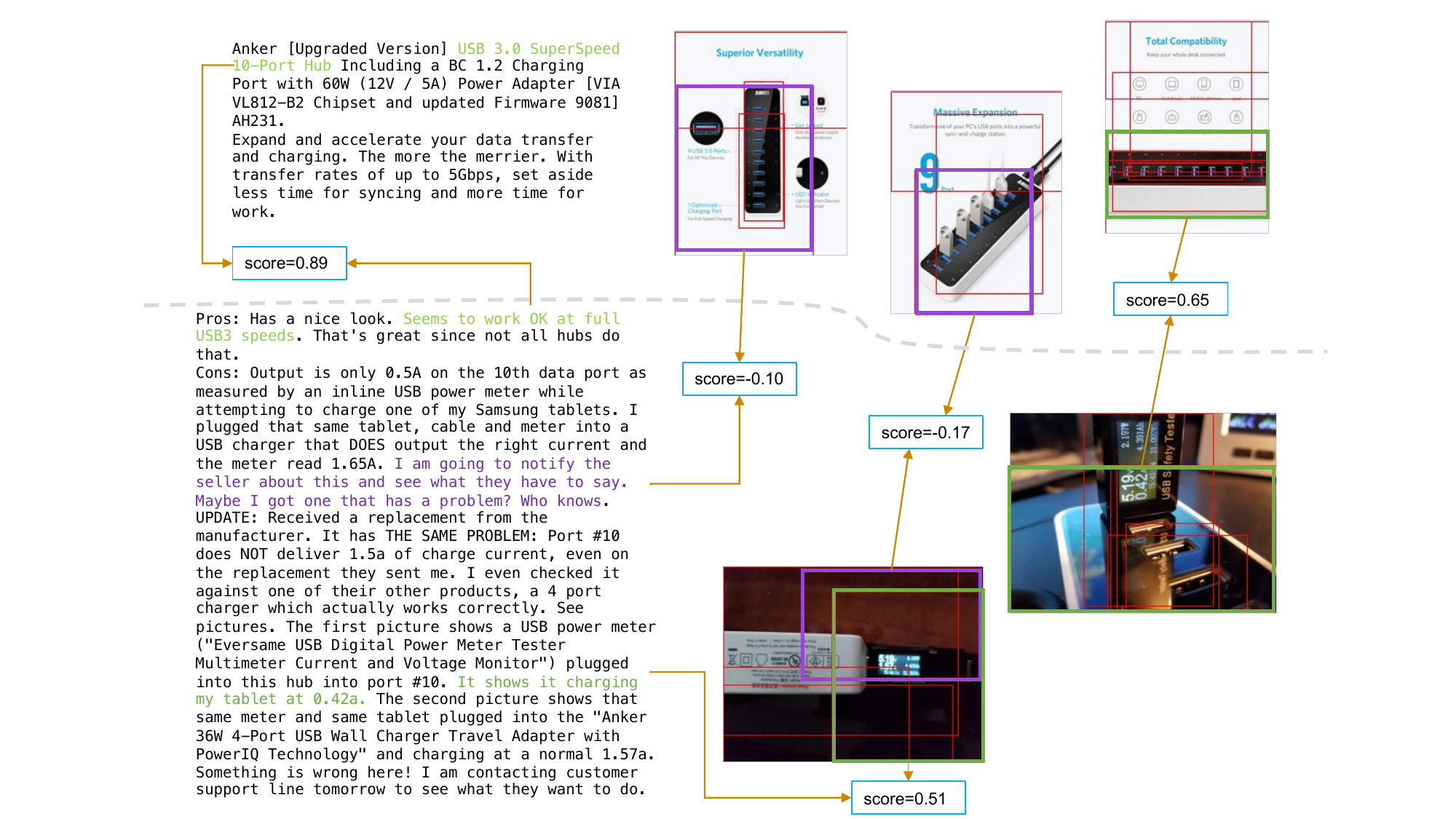}
    \caption{A case study from Amazon-MRHP dataset. 
    The upper and lower part of the figure is the product and review post respectively.
    Green and purple are instance pairs that produce high and low scores that are selected/not selected into the final feature vector. 
    For the matching of n-gram words, we display the largest matching scores between individual words in that scope and the other elements.}
    \label{fig:case_study}
\end{figure*}

To further understand the model's functional mechanism, we randomly pick an example from the Amazon--electronics and visualize some matching scores during the test time in Figure~\ref{fig:case_study}.
There are several valuable points to underline.
First, when the model achieves the best performance, its matching scores can reflect the correlation between some semantically matched feature pairs.
For instance, the RoI-RoI matching score of -0.17 is produced by the two RoIs that enclose different objects in their respective images, and thus the correlation between them is very weak, and a near 0 score is obtained, while the two boxes that contain the port hub achieve a relatively high matching score.
Second, text-text matching may act as word matching. 
It is hard to attribute the 0.89 matching score of those two sentences to the high semantic similarity between those two text snippets since their semantic meanings are different, only to share some common words.
These two discoveries reveal that~\modelname~attends to more than semantic matching, and just a certain number of correct matching scores could make up the last features for its accurate prediction, in accord with the conclusion about $K$ values. 

\subsection{Comparison on examples with GBDT}
We further randomly draw two examples which PREMISE gives accurate predictions but GBDT, the state-of-the-art baseline, fails. 
The original review context (including text and attached image) together with the predictions from GBDT and our model towards these two examples are displayed  in~\Cref{tab:elec_product}~to~\Cref{tab:case_study2}. 

We find that~\modelname~ ranks these reviews in correct order (it has the same ranking sequence $1\to 2 \to 3$  and $1 \to 2 \to 3 \to 4$ as the ground truth's in the two examples respectively) even trained and tested with normalized score whose values range from 0 to 1 (different from the annotated scores $s\in [0,4]$). In the given examples, GBDT flips the order of 'B005NGQWL2-9' and 'B005NGQWL2-65' in example 1 and the order of 'B00H4O1L9Y-111' and 'B00H4O1L9Y-122' in example 2.
This could imply that matching-based modeling can make more accurate predictions than fusion-based modeling.

\begin{table*}[ht]
    \resizebox{\textwidth}{!}{
        \begin{tabular}{p{0.2\linewidth}|p{0.7\linewidth}}
        \toprule
           Product ID & Introduction \\
         \midrule
            B005NGQWL2 & Expand and accelerate your data transfer and charging. <br><br><b>The more the merrier.</b><br>With transfer rates of up to 5Gbps, set aside less time for syncing and more time for work. With 10 data terminals to choose from, forget about ever having to switch or unplug again.<br><br><b>Fast charging.</b><br>10th-port dual functionality enables fast charges of up to 1.5 amps with BC 1.2 charging-compliant devices, while simultaneously transferring data. Charge via a power adapter for higher 2 amp speeds with all USB-enabled devices when hub is disconnected from an active USB port, or your computer is off or in sleep mode. Dual functions, duly facilitated.<br><br><b>A mainstay for the future.</b><br>Featuring a high-grade chipset and a powerful 60W adapter, this hub ensures a stable power supply while you work. Get steady operation for years to come. Whether at home or in the office, add another can't-do-without fixture to your desk.<br><br><b>BC 1.2 Charging-Compliant Devices:</b><br>Apple: iPhone 5 / 5s, iPad Air, iPad mini / mini 2<br>Samsung: Galaxy S3 / S4, Galaxy Note 1 / 2, Galaxy Mega, Galaxy Mini, Exhilarate, Galaxy Tab 2 10.1<br>Google: Nexus 4 / 5 / 7 / 10<br>Sony: Xperia TX <br>Nokia: Lumia 920, Lumia 1020 <br><br><b>System Requirements</b><br>Windows (32/64 bit) 10 / 8.1 / 8 / 7 / Vista / XP, Mac OS X 10.6-10.9, Linux 2.6.14 or later.<br>Mac OS X Lion 10.7.4 users should upgrade to Mountain Lion 10.8.2 or later to avoid unstable connections.<br><br><b>Compatibility</b><br>2.4GHz wireless devices, MIDI devices and some USB 3.0 devices may not be supported. Try using the host port or a USB 2.0 connection.<br><br><b>Power Usage</b><br>For a stable connection, don't use this hub with high power-consumption devices, such as external hard drives. The hub will sync but not charge tablets and other devices which require a higher power input.

        \subfigure{\includegraphics[height=\csfigheight]{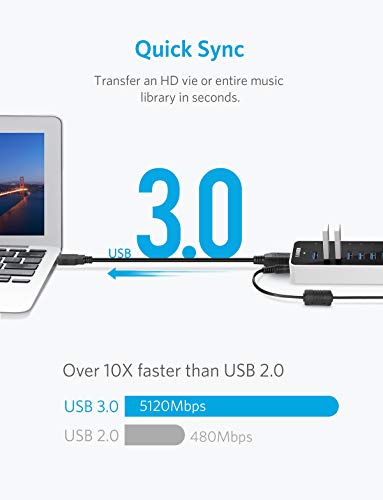}}
        \subfigure{\includegraphics[height=\csfigheight]{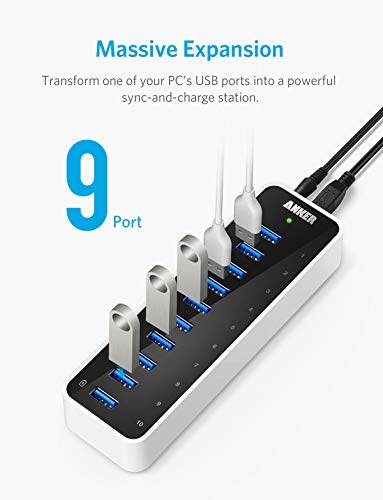}}
        \subfigure{\includegraphics[height=\csfigheight]{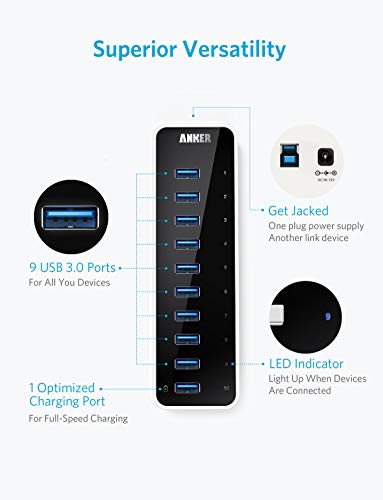}}
        \subfigure{\includegraphics[height=\csfigheight]{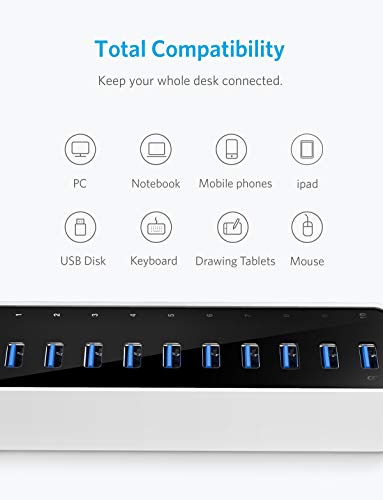}}
        \subfigure{\includegraphics[height=\csfigheight]{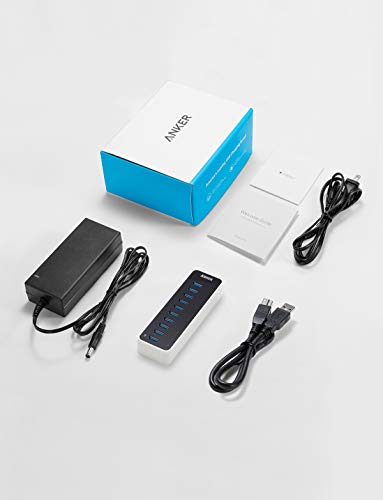}} 
        \subfigure{\includegraphics[height=\csfigheight]{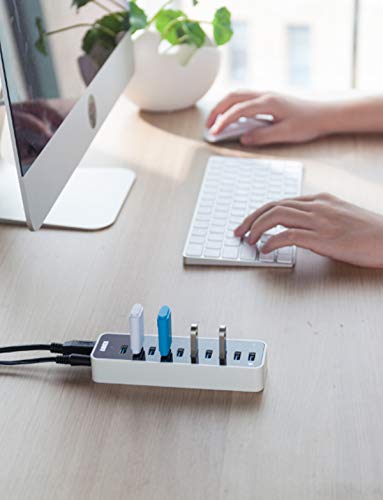}} 
        \\
        \bottomrule
        \end{tabular}
    }
    \caption{(Example 1 of 2) Product introduction of an example from Amazon Electronics dataset. Some special characters have been removed for better readability.}
    \label{tab:elec_product}
\end{table*}

\begin{table*}[ht!]
    \resizebox{\textwidth}{!}{
        \begin{tabular}{p{0.2\linewidth}|p{0.7\linewidth}|p{0.07\linewidth}|p{0.07\linewidth}|p{0.07\linewidth}}
        \toprule
        Review ID & Content & GT & GBDT & Ours \\
        \midrule
        B005NGQWL2-14 & 
        Pros:
        
        Has a nice look. Seems
        to work OK at full USB3 speeds. That's great since not all hubs do that.
        
        Cons: 

        Output is only 0.5A on the 10th data port as measured by an inline USB power meter while attempting to charge one of my Samsung tablets. I plugged that same tablet, cable and meter into a USB charger that DOES output the right current and the meter read 1.65A. 
        
        I am going to notify the seller about this and see what they have to say. Maybe I got one that has a problem? Who knows. 
        
        UPDATE: Received a replacement from the manufacturer. It has THE SAME PROBLEM: Port \#10 does NOT deliver 1.5A of charge current, even on the replacement they sent me. I even checked it against one of their other products, a 4 port charger which actually works correctly. See pictures. The first picture shows a USB power meter (``Eversame USB Digital Power Meter Tester Multimeter Current and Voltage Monitor") plugged into this hub into port \#10. It shows it charging my tablet at 0.42a. The second picture shows that same meter and same tablet plugged into the "Anker 36W 4-Port USB Wall Charger Travel Adapter with PowerIQ Technology" and charging at a normal 1.57a. Something is wrong here! I am contacting customer support line tomorrow to see what they want to do.
        
        \subfigure{\includegraphics[height=\csfigheight]{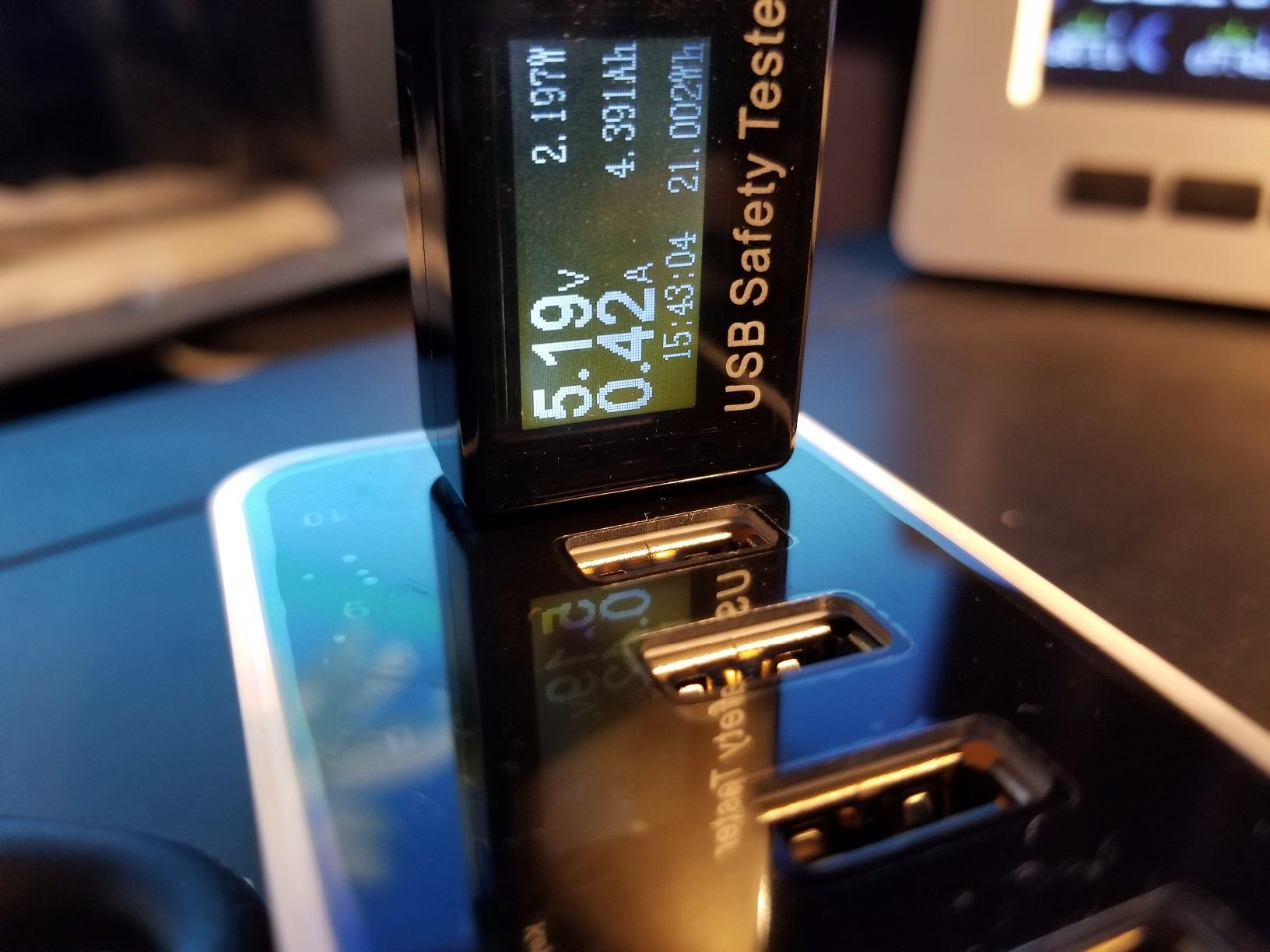}}
        \subfigure{\includegraphics[height=\csfigheight]{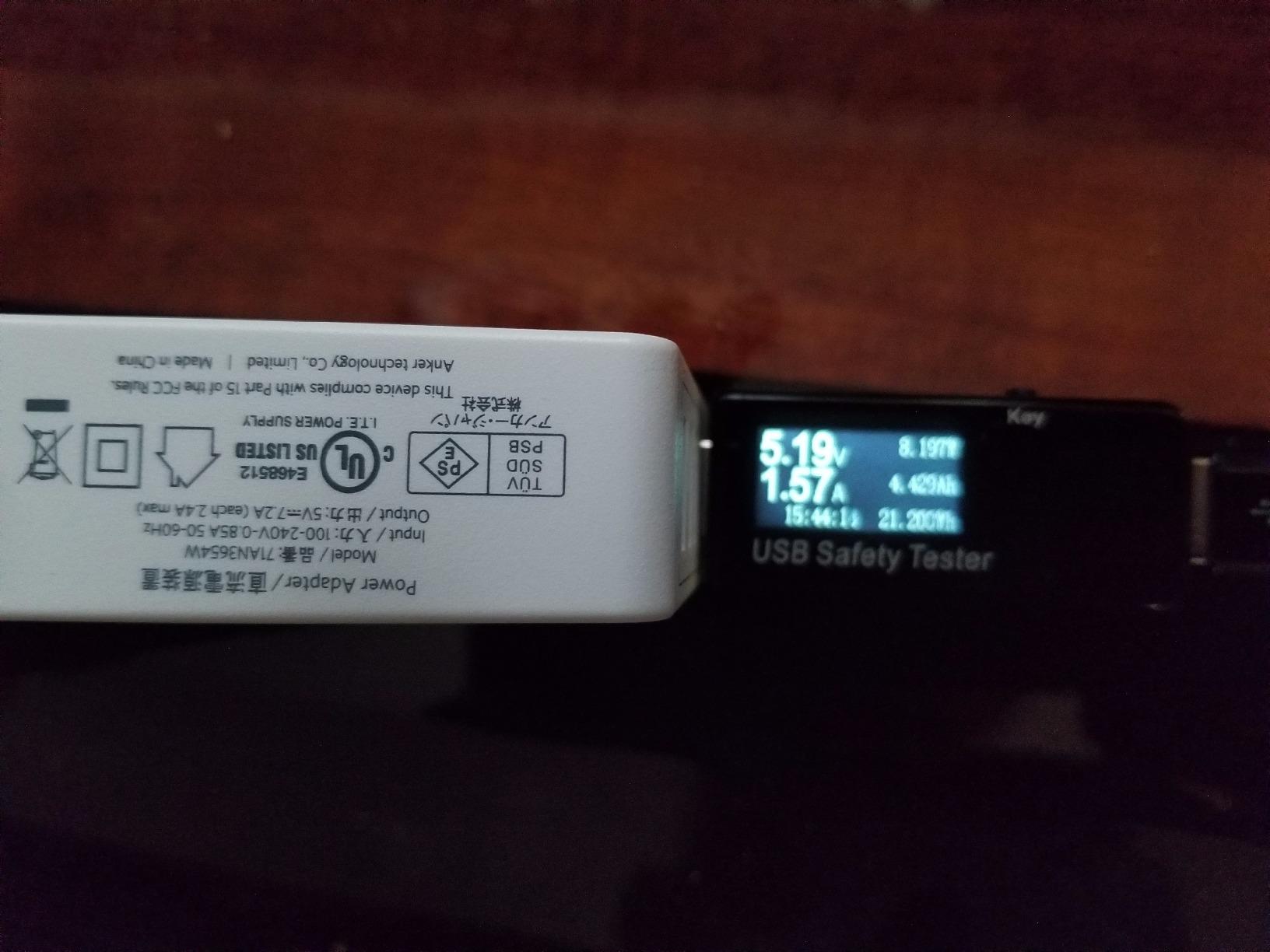}}
        & 3.00 & 0.86 & 0.89 \\
        \midrule
        B005NGQWL2-9 & I rarely write a negative review, in fact almost never. This AnkerDirect 10-Port USB Data hub lasted only about 2 months. Now none of the USB ports work.  For the first couple of weeks all the ports seemed fine. Then one by one they stopped working. Power gets to the unit and the USB ports on my MAC work fine. I've been waiting for a replacement from the company ever since April 20th 2017, after sending their support group my address as requested,  but it had not arrived.
        
        I wish to amend this review by saying that the Anker customer service folks were very helpful in rectifying this situation. After some checks on the original item at their direction, the Anker folks came to the conclusion that I had a defective product and quickly replaced it with a new model. I've had a couple of days to test it out and it appears to be working just fine. I have always felt that a product or service can go bad but it is the company's response to that problem, should it arise, that gains my respect and future business.

        \subfigure{\includegraphics[height=\csfigheight]{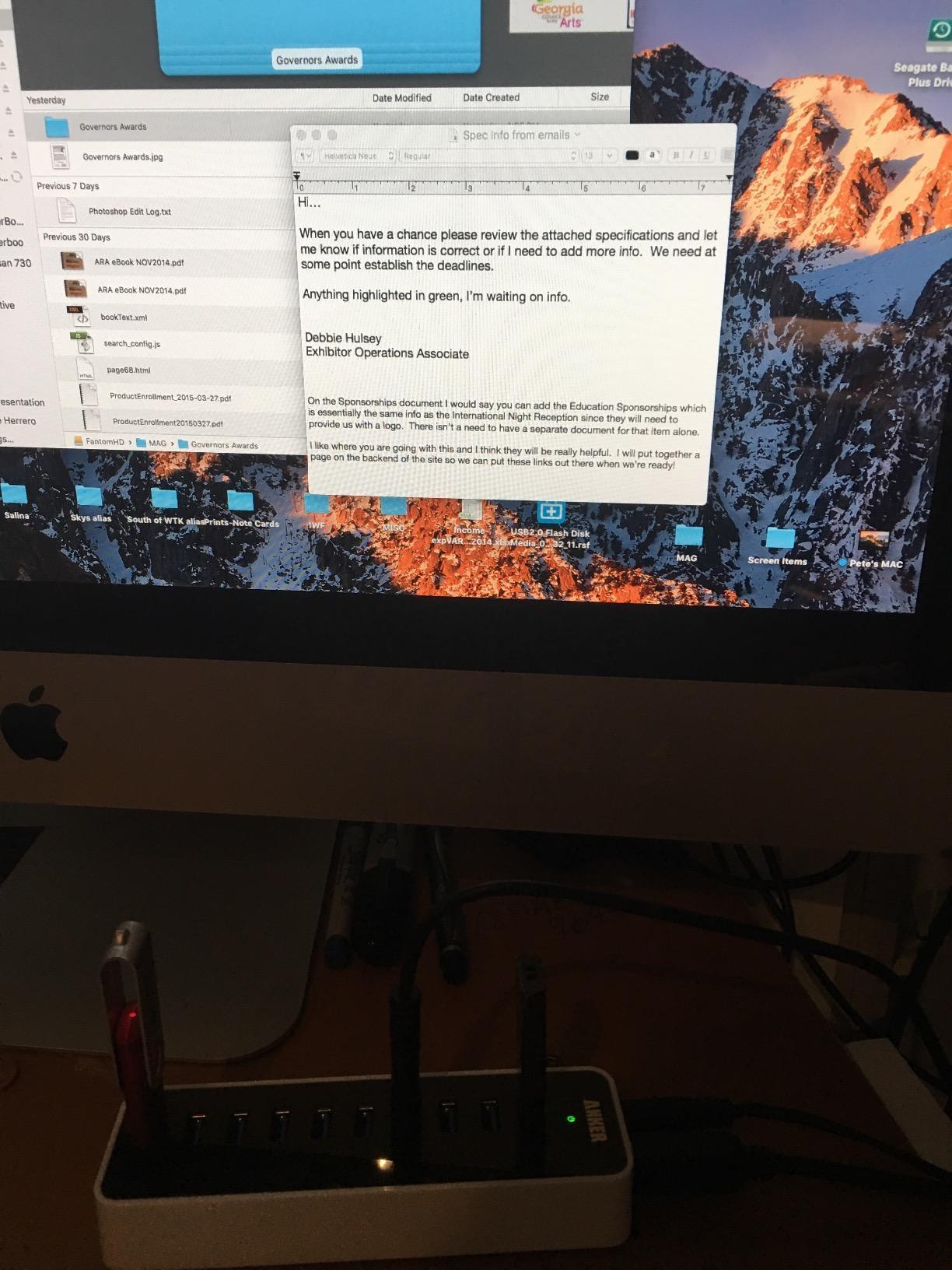}}
        & 2.00 & 0.64 & 0.52 \\
        
        \midrule
        B005NGQWL2-65 & Works very great, powers all of my USB connections, I have a Asus Gaming Laptop which only has 4 USB ports and I needed to have a blue yeti mic, a Logitech Webcam c920,  razer keyboard chroma,  2tb hard drive,  and a Xbox one controller wireless adapter connected to it. 
        
        So far nothing has disconnected or malfunctioned. I would definitely recommend this to my friends and familiy. 
        
        \subfigure{\includegraphics[height=\csfigheight]{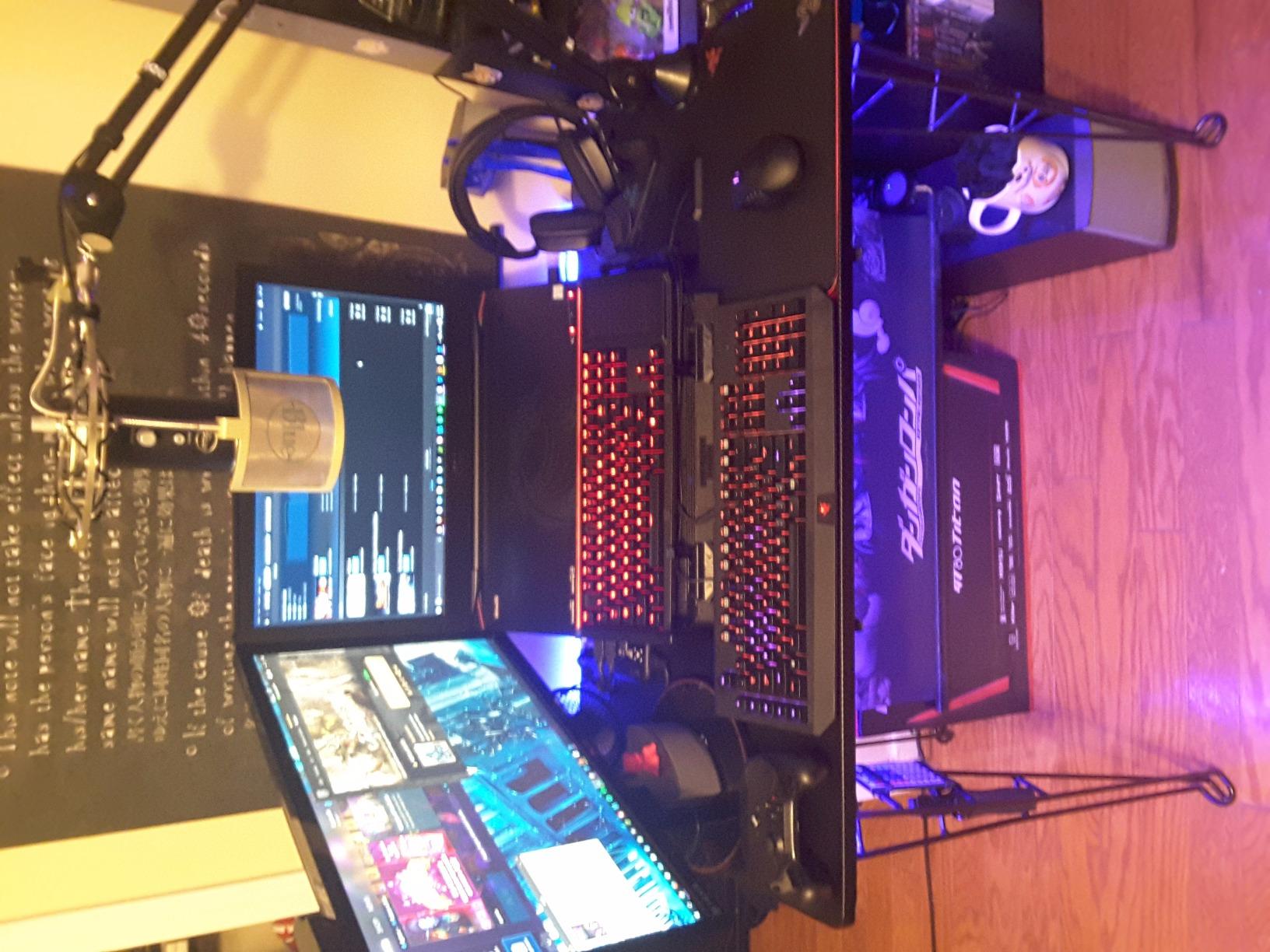}}
        \subfigure{\includegraphics[height=\csfigheight]{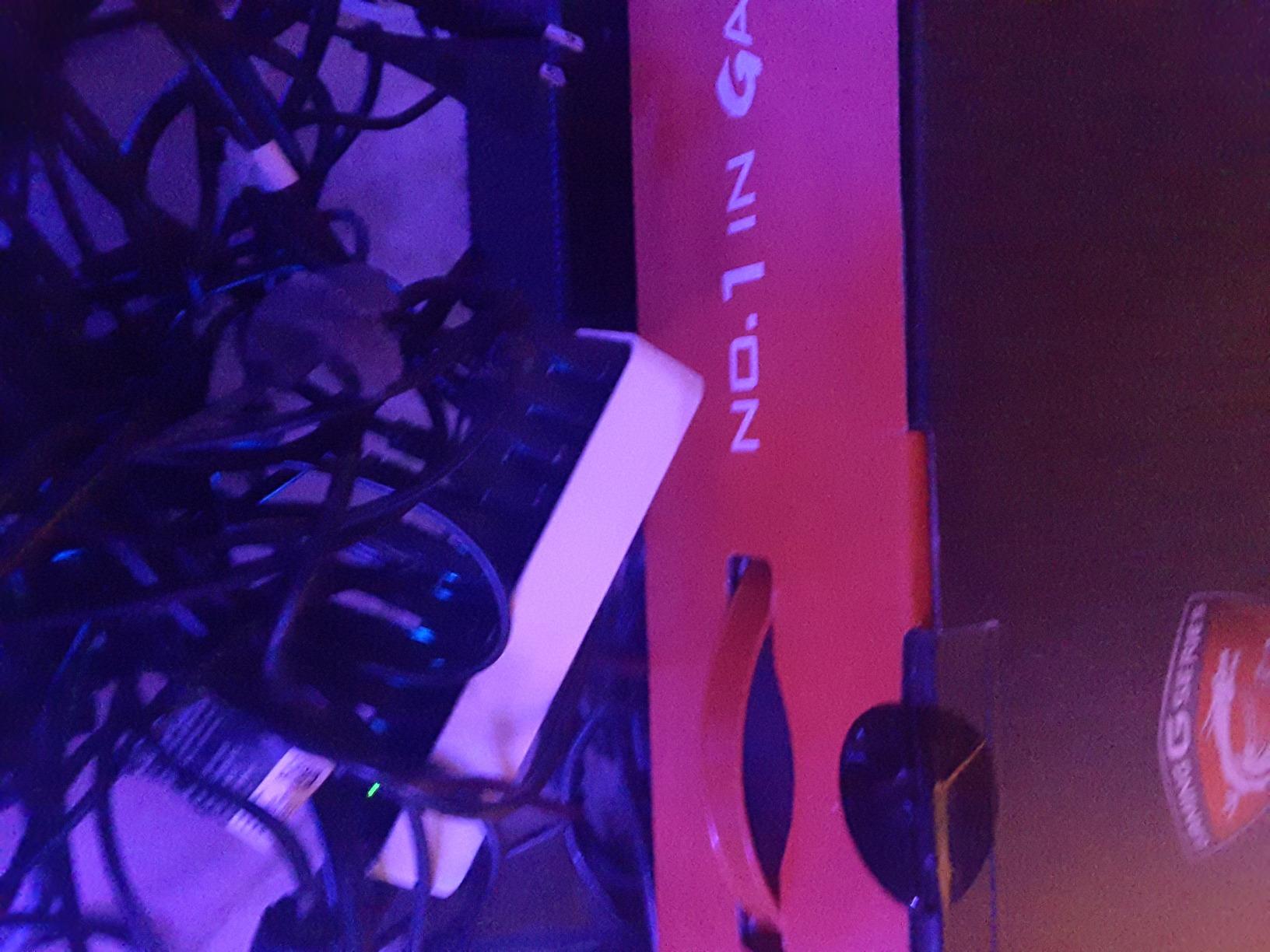}} & 1.00 & 0.75 & 0.31 \\
        \bottomrule
        \end{tabular}
    }
    \caption{(Example 1 of 2) Comparison between our model and GBDT on an example from Amazon electronics dataset. The ground truth (GT) scores are annotated ones, while the scores below GBDT and ours are normalized ones.}
    \label{tab:case_study1}
\end{table*}

\begin{table*}[ht!]
    \resizebox{\textwidth}{!}{
        \begin{tabular}{p{0.2\linewidth}|p{0.7\linewidth}}
        \toprule
           Product ID & Introduction \\
         \midrule
            B00H4O1L9Y & Cook food to perfection with the T-fal OptiGrill GC702D53 electric indoor grill. This indoor grill offers versatility and convenience for any grilled meals. Choose from six pre-set programs: Burger, Poultry, Sandwich, Sausage, Red Meat, and Fish. The grills precision grilling technology with sensors measures the thickness of food for auto cooking based on the program selected. When the flashing light turns solid purple, the grill has properly preheatedplace food on the grill, lower the lid, and it takes care of the rest. A cooking-level indicator light changes from yellow to orange to red signifying the cooking progress with audible beeps that alert when food gets to each stage: rare, medium, and well-done. Take food off the grill once its reached your preferred level of doneness. Along with the six pre-set programs, the electric grill provides two additional cooking options: Frozen mode for defrosting and fully cooking frozen food and Manual mode for cooking vegetables or personal recipes. (Note: when preheating for a pre-set program, keep the lid closed or the grill will automatically switch to Manual mode.) The OptiGrill features a powerful 1800-watt heating element, user-friendly controls ergonomically located on the handle, and die-cast aluminum plates with a nonstick coating for effortless food release. The slightly angled cooking plates allow fat to run away from food and into the drip tray for healthier results, and the drip tray and cooking plates are removable and dishwasher-safe for quick cleanup. Housed in brushed stainless steel, the OptiGrill electric indoor grill makes an attractive addition to any counter.  
    
            \subfigure{\includegraphics[height=\csfigheight]{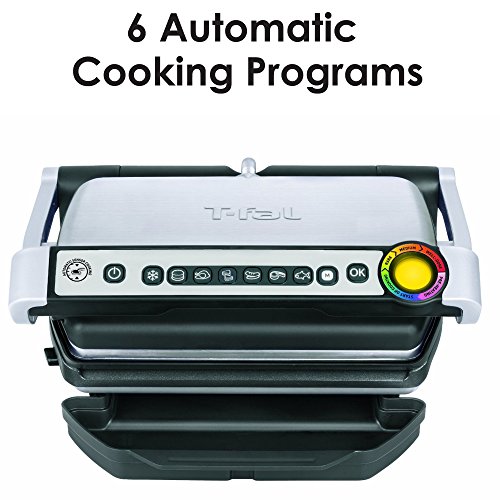}}
            \subfigure{\includegraphics[height=\csfigheight]{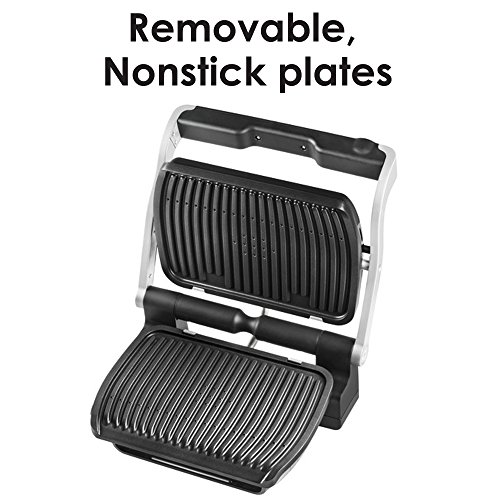}}
            \subfigure{\includegraphics[height=\csfigheight]{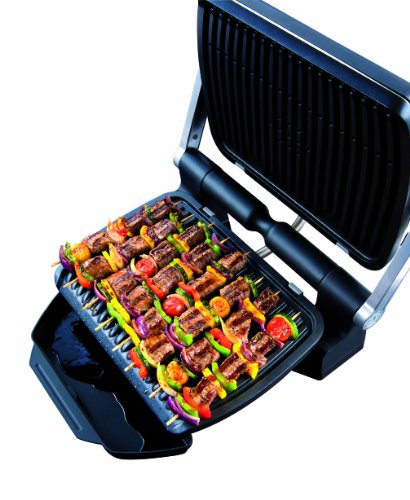}}
            \subfigure{\includegraphics[height=\csfigheight]{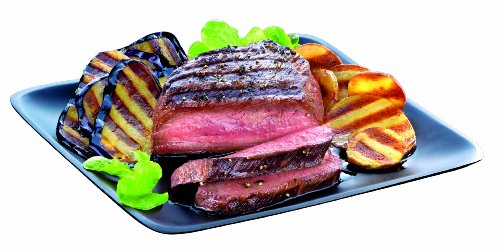}} \\
        \bottomrule
        \end{tabular}
    }
    \caption{(Example 2 of 2) Product introduction of an example from Amazon Home dataset.}
    \label{tab:home_product}
\end{table*}

\begin{table*}[ht!]
    \resizebox{\textwidth}{!}{
        \begin{tabular}{p{0.2\linewidth}|p{0.7\linewidth}|p{0.07\linewidth}|p{0.07\linewidth}|p{0.07\linewidth}}
        \toprule
        Review ID & Content & GT & GBDT & Ours \\
        \midrule
        B00H4O1L9Y-111 & 
        I want to preface this by saying that I always prefer food grilled on our big outdoor propane grill.  It's just a superior method of cooking.  That being said, if you don't have an outdoor grill, or even if you do and are sometimes unable to cook with it due to lack of time, running out of propane, inclement weather, laziness...  then this is a FABULOUS option to still get the grilled food you loved SUPER FAST and SUPER EASY!!  We got 5 (yes FIVE) George Foreman grills for our wedding.  I re-gifted 4 of them and kept one and have used it off and on for a long time, but every time I have to clean it afterward I swear I'm never going to use it again because it's such a pain and it never quite gets clean, especially in the area where the hinges are.  That problem is no more....
        
        \subfigure{\includegraphics[height=\csfigheight]{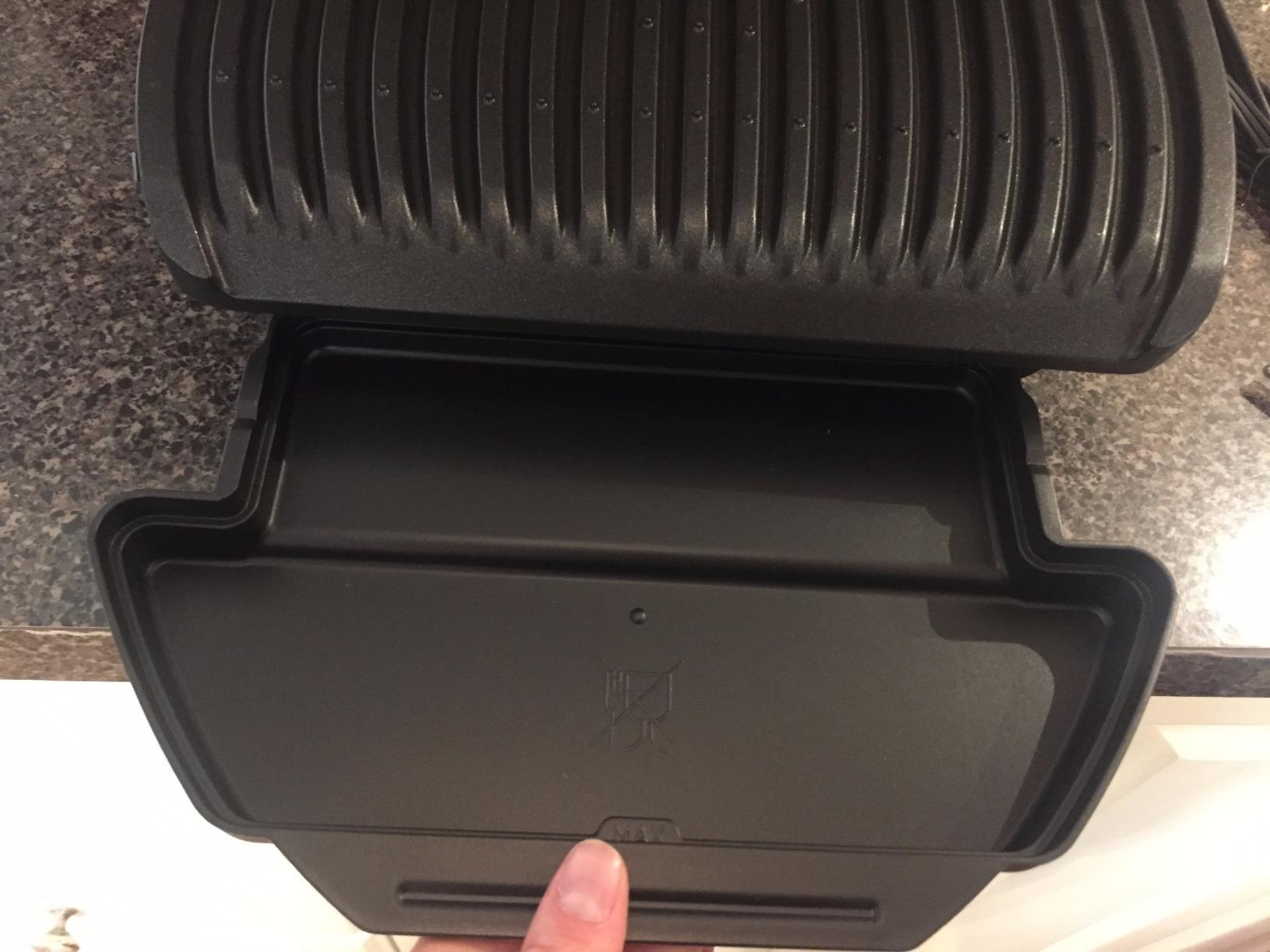}}
        \subfigure{\includegraphics[height=\csfigheight]{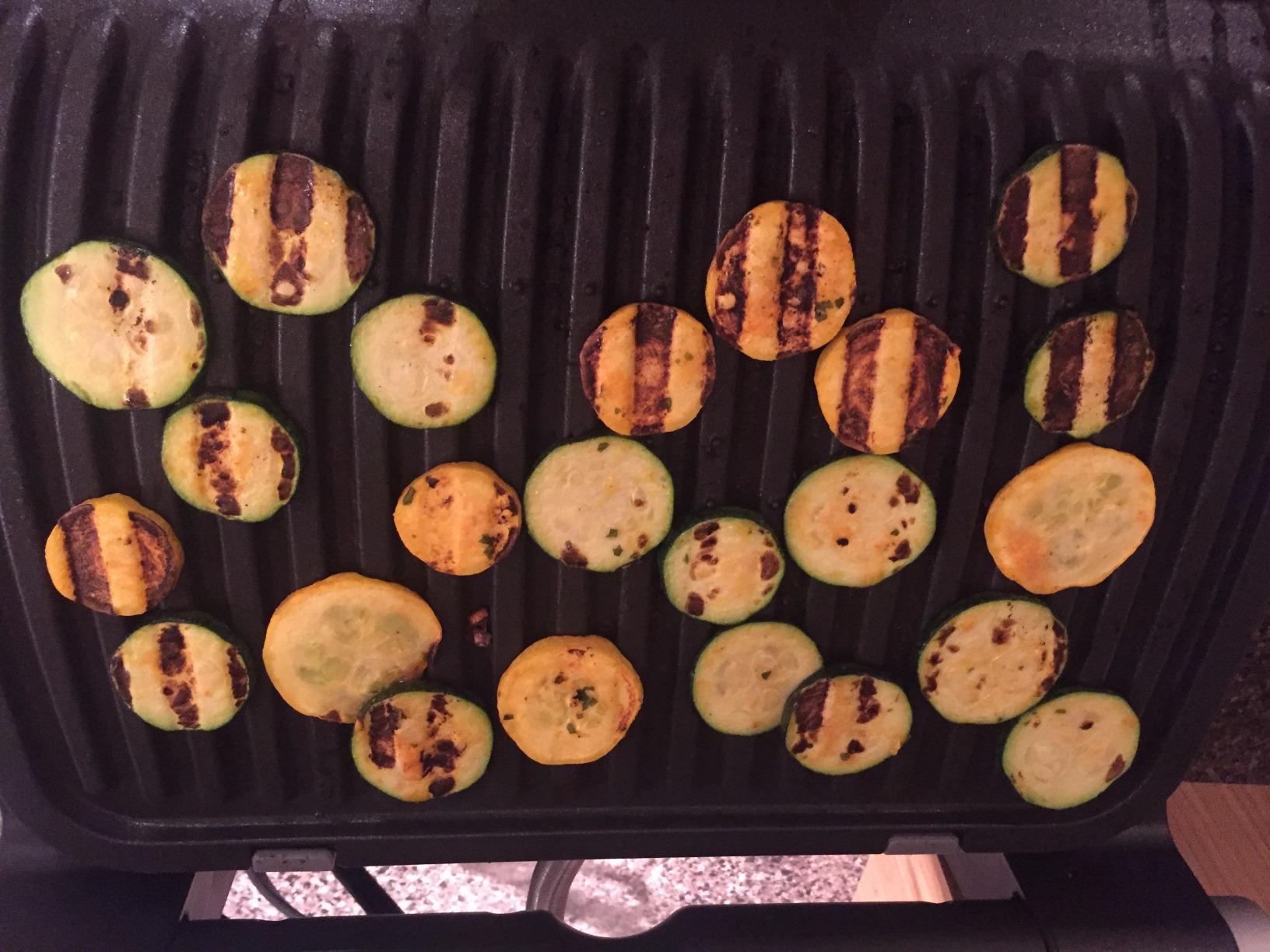}}
        \subfigure{\includegraphics[height=\csfigheight]{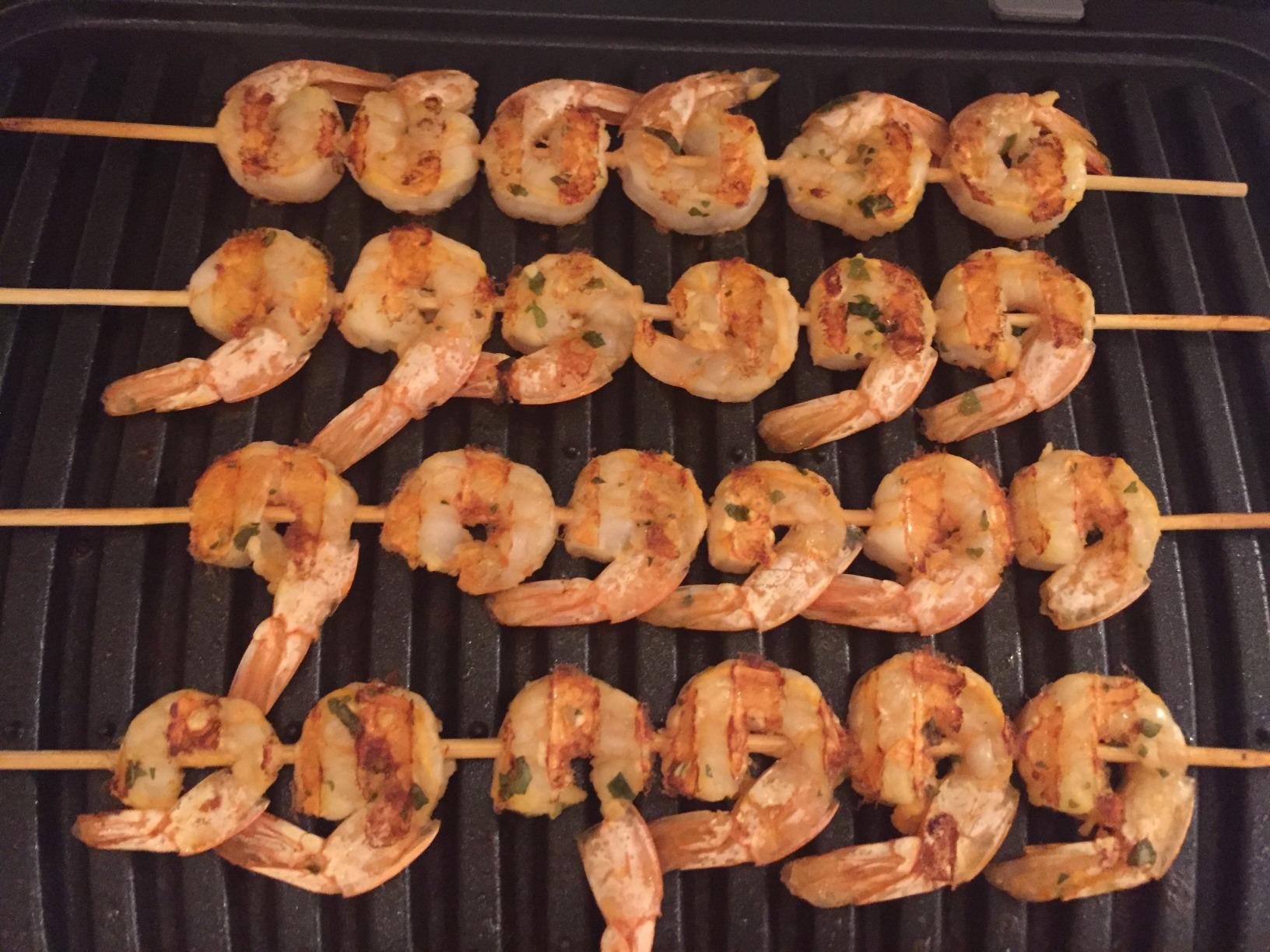}}
        \subfigure{\includegraphics[height=\csfigheight]{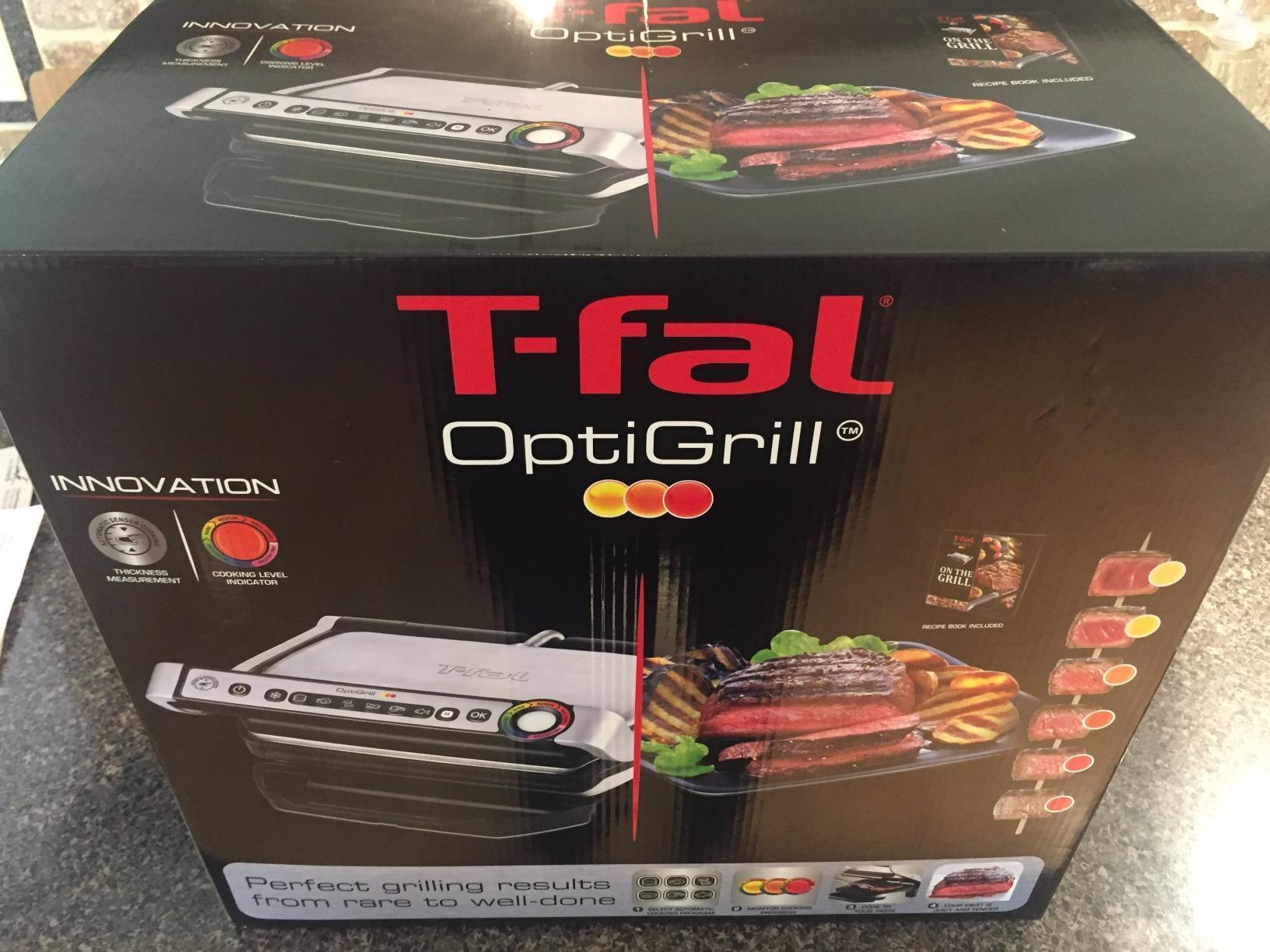}}
        \subfigure{\includegraphics[height=\csfigheight]{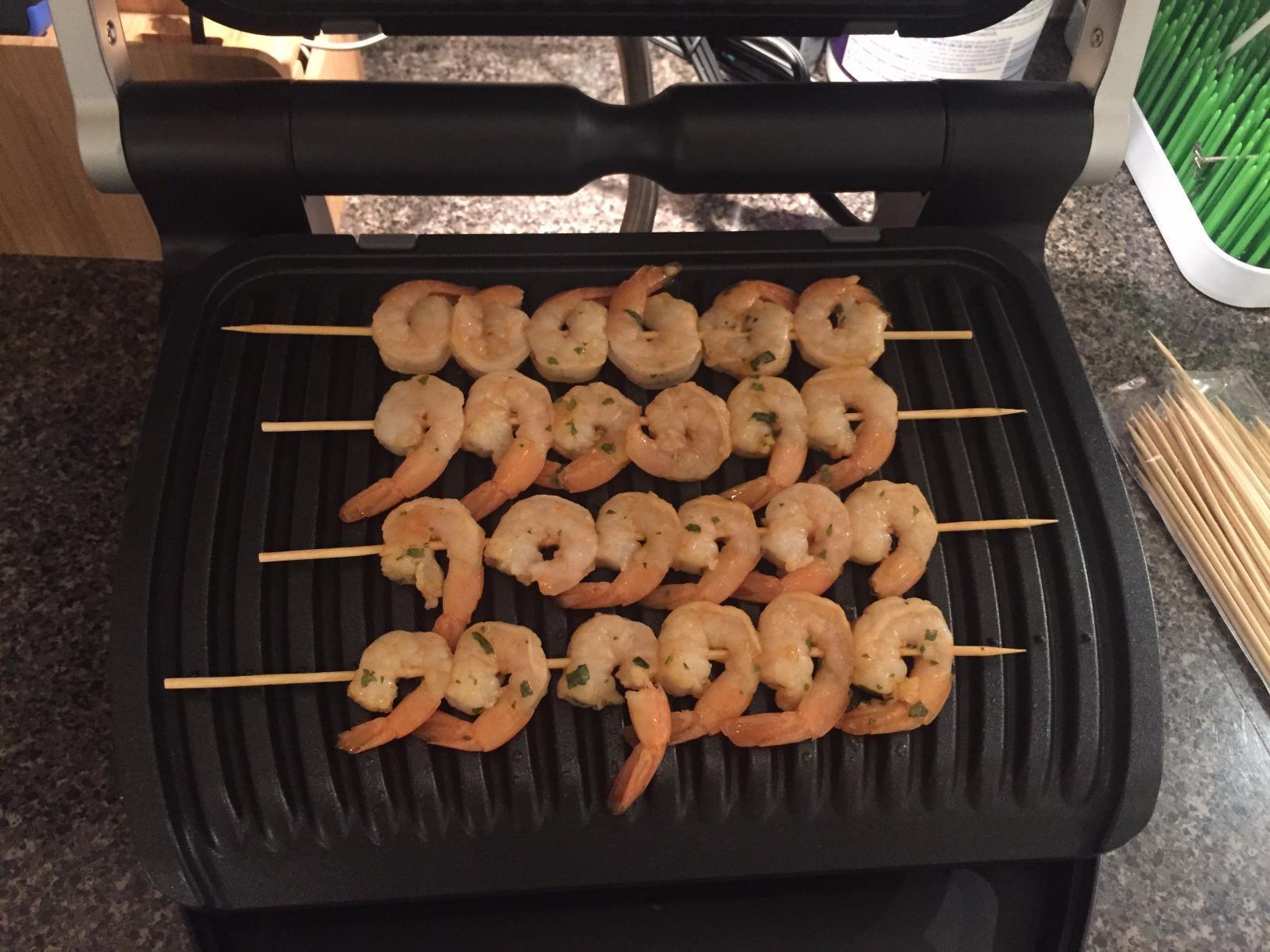}}
        \subfigure{\includegraphics[height=\csfigheight]{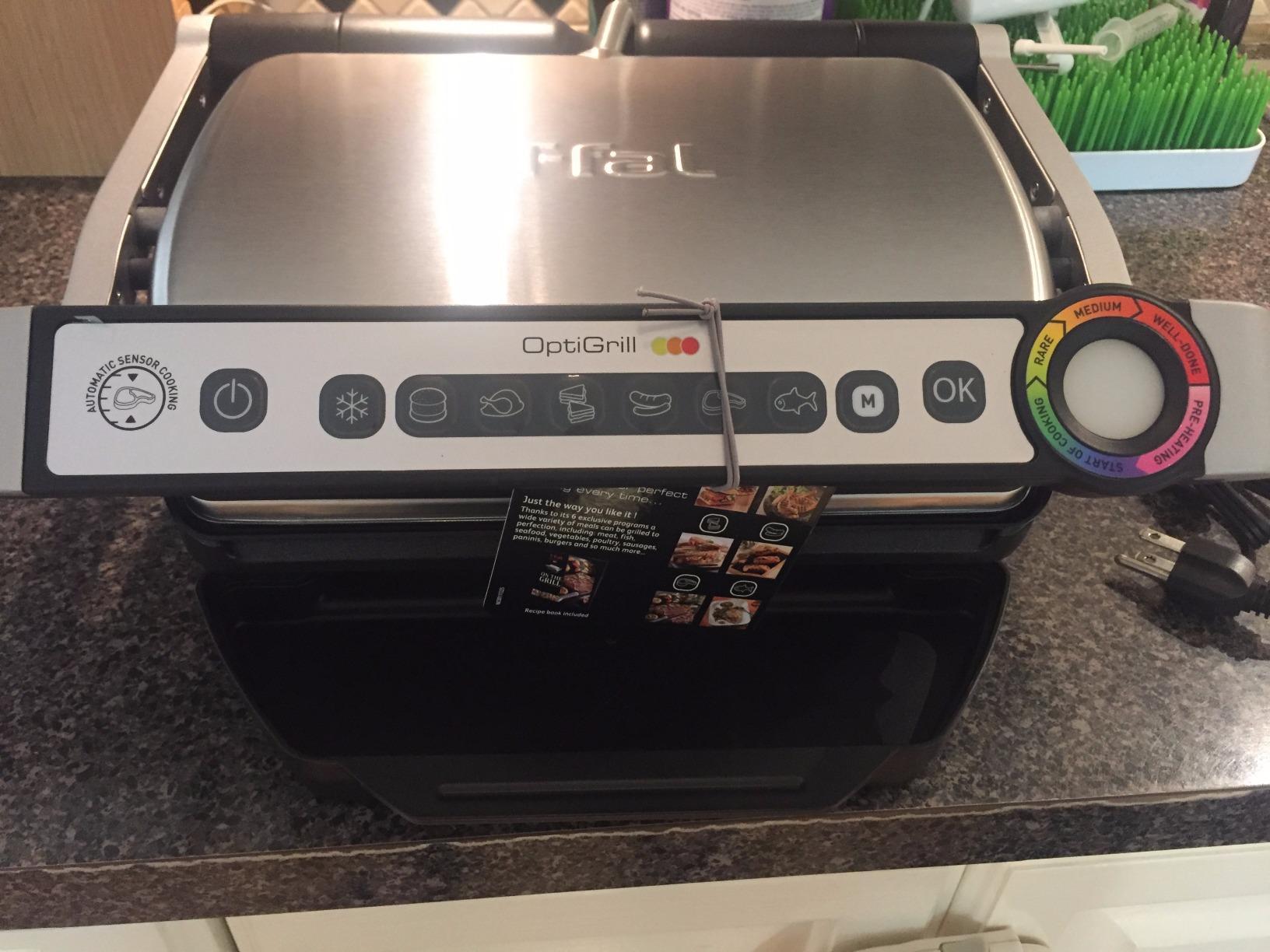}}
        \subfigure{\includegraphics[height=\csfigheight]{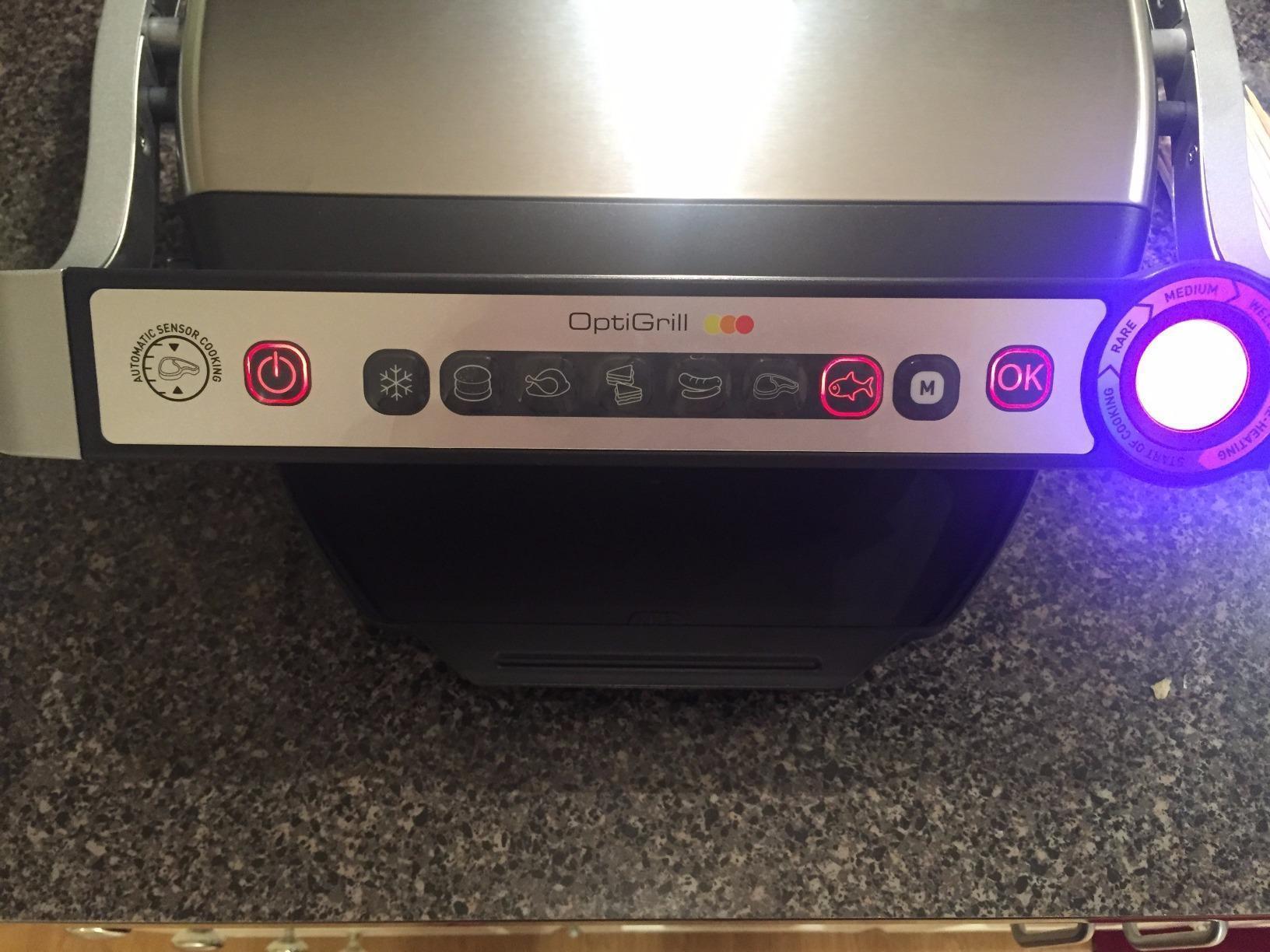}}
        & 4.00 & 0.58 & 0.71 \\
        \midrule
        B00H4O1L9Y-122   &
        My mom got this on her account. I thought she was crazy to spend so much on what looked like a glorified George Foreman grill but I was wrong, this thing is the bomb. Here is what I like about it;-Heats up super quick. I remember my old George Forman grill took a lot longer.-The presets for the type of food you are cooking must be working because nothing turns out overcooked.
        
        The nonstick removable plates. So far, I haven't had any food stick to the plates and I don't use spray or oil. Being able to take them off and wash them in the sink or dishwasher is by far the best part, I used to hate wasting a million paper towels and burning my hands on my old foreman grill and still didn't feel like it was clean. 
        
        Doesn't create a lot of smoke. When I used to use my old foreman grill I was always setting off the smoke alarm, this grill doesn't do that.
        
        \subfigure{\includegraphics[height=\csfigheight]{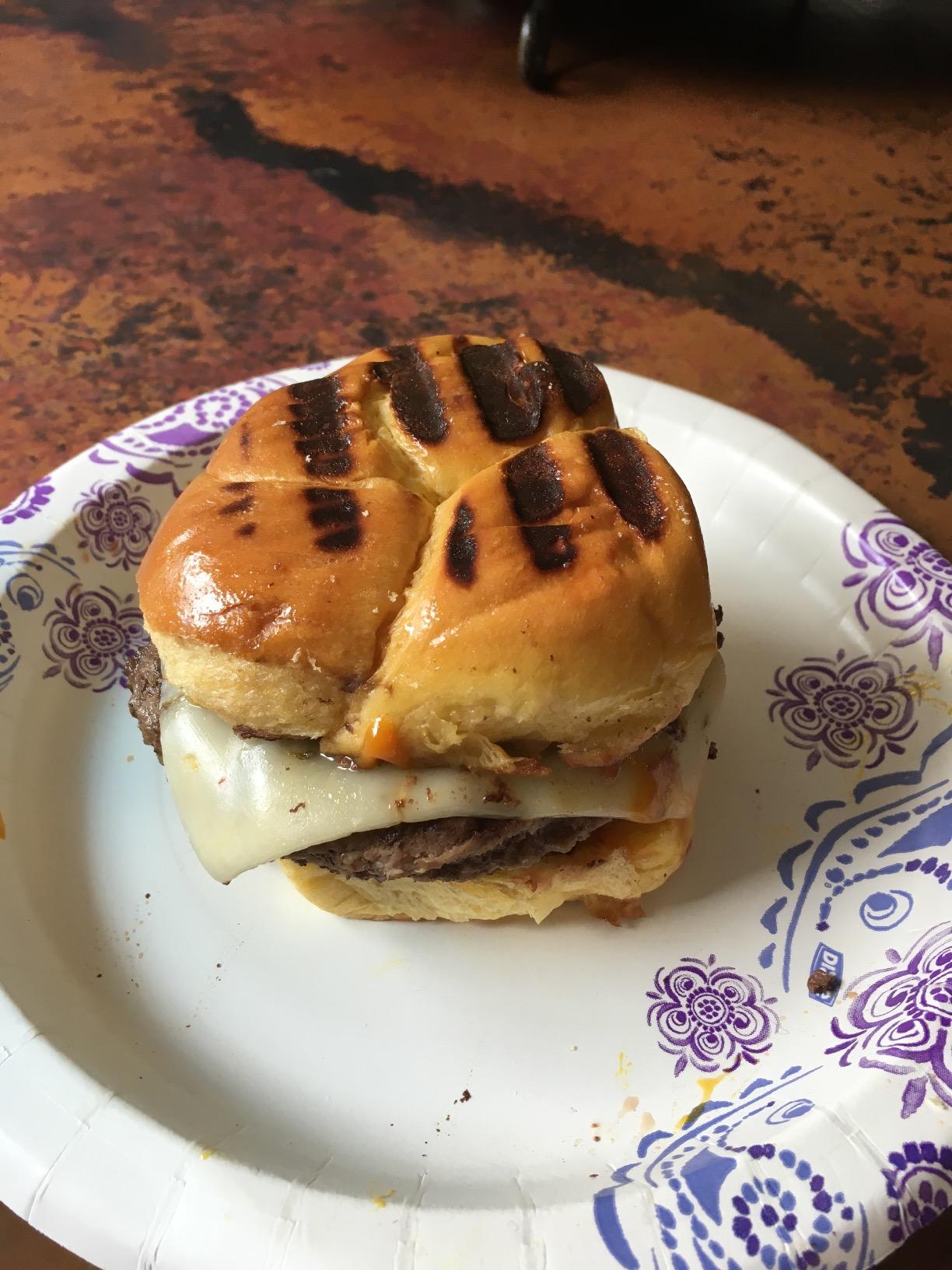}}
        \subfigure{\includegraphics[height=\csfigheight]{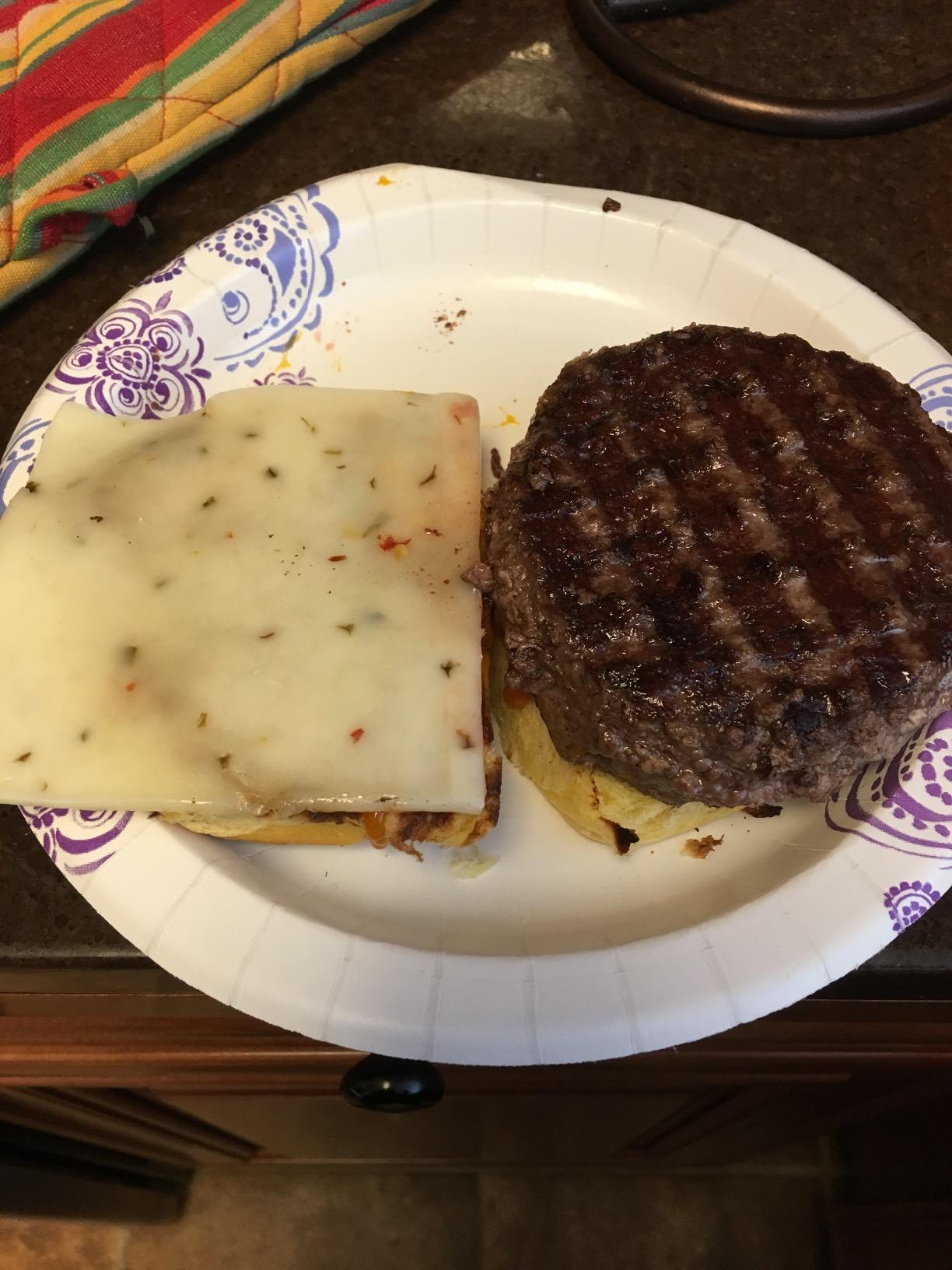}}
        \subfigure{\includegraphics[height=\csfigheight]{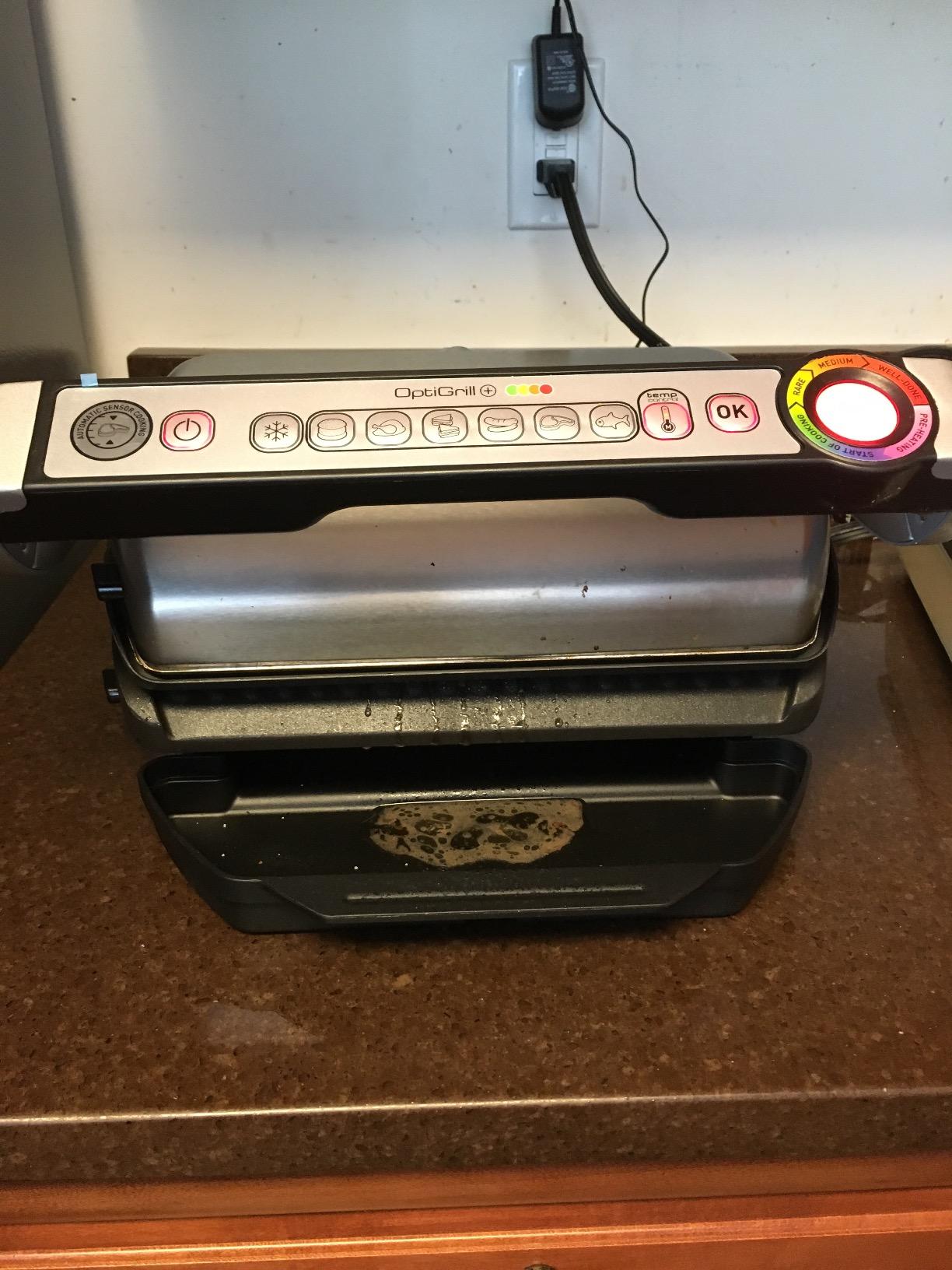}}
        \subfigure{\includegraphics[height=\csfigheight]{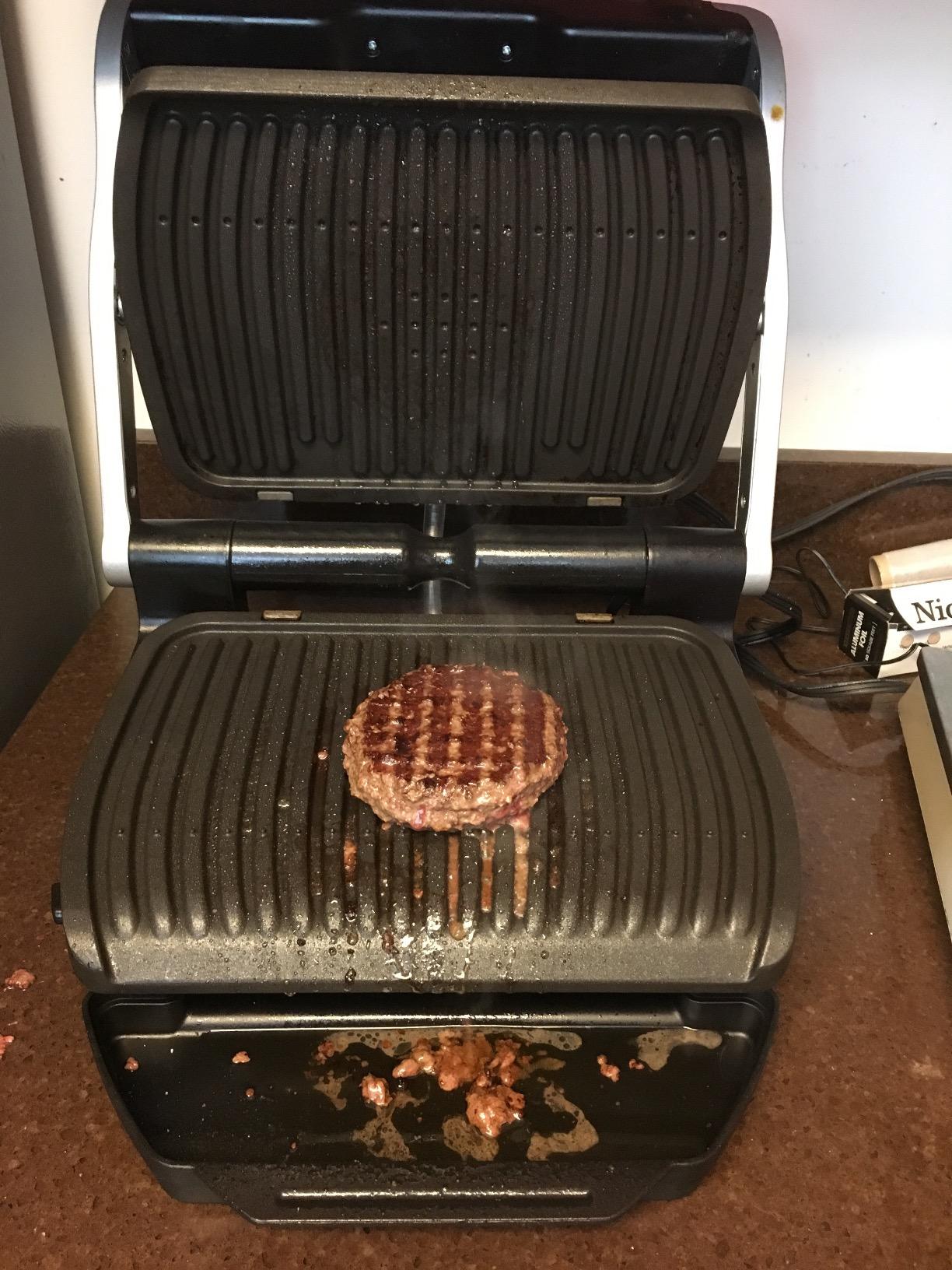}}
        & 3.00 & 0.65 & 0.62 \\
        
        \midrule
        B00H4O1L9Y-148 & I have used this grill 4 times now and everything I've cooked has turned out amazing! I am so impressed with this grill. It is real quick to preheat and has cooked everything perfectly so far from burgers to chicken sausages to kabobs. We haven't tried anything from the cookbook included but we definitely want to. One of the best things is that the plates are detachable and dishwasher safe! Super easy cleanup.
        
        \subfigure{\includegraphics[height=\csfigheight]{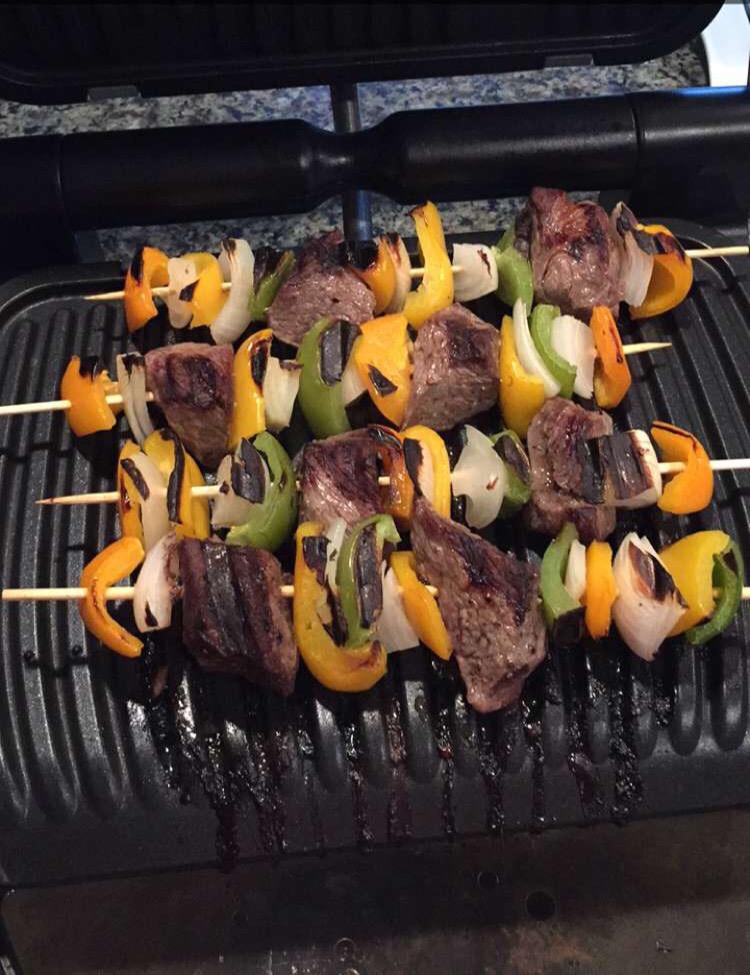}}
        \subfigure{\includegraphics[height=\csfigheight]{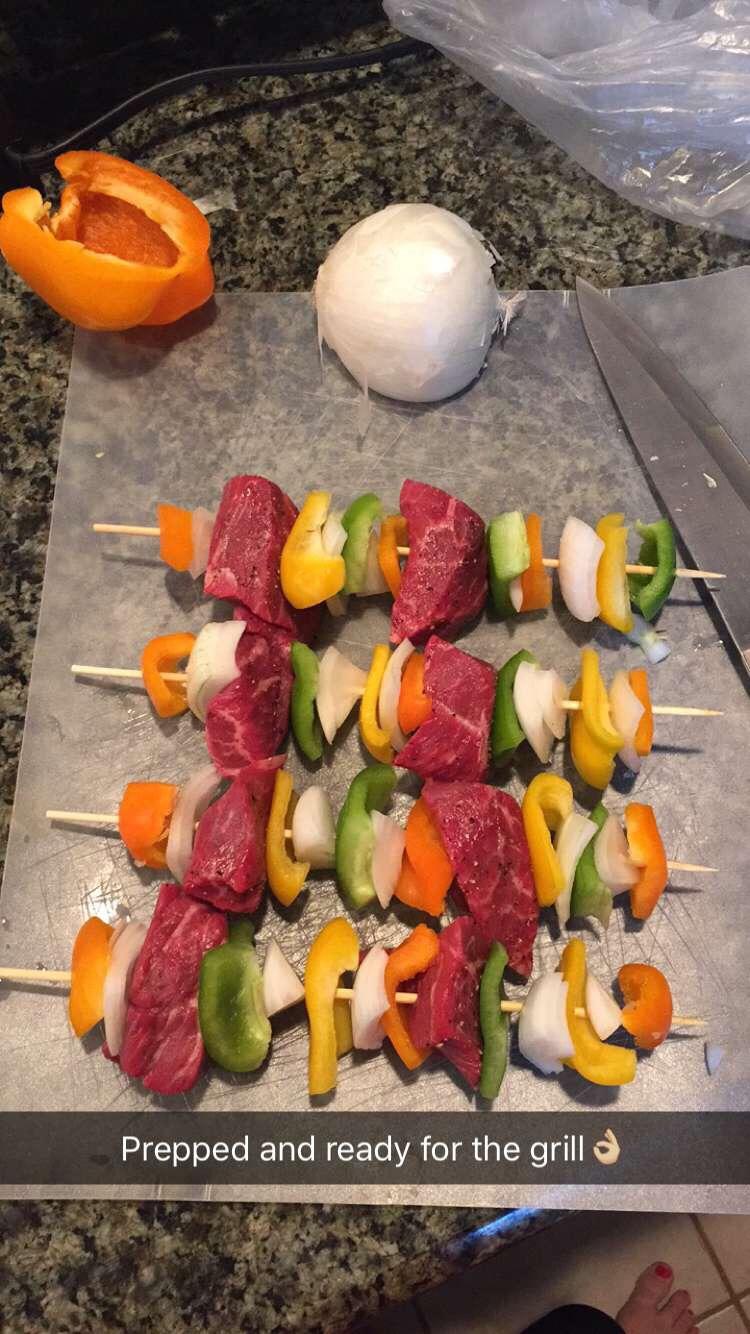}} & 2.00 & 0.49 & 0.42 \\
        \midrule
        B00H4O1L9Y-59 & One of the best tools for preparing clean food (if that's what you choose). Cooks in minutes and cleans just as fast. Gives that grill experience within a compact structure. Definitely saves time, I use this thing at least ounce a day, on prep days 3-5 times. If you want to loose wait; it starts in YOUR kitchen by preparing your meals. 
        
        \subfigure{\includegraphics[height=\csfigheight]{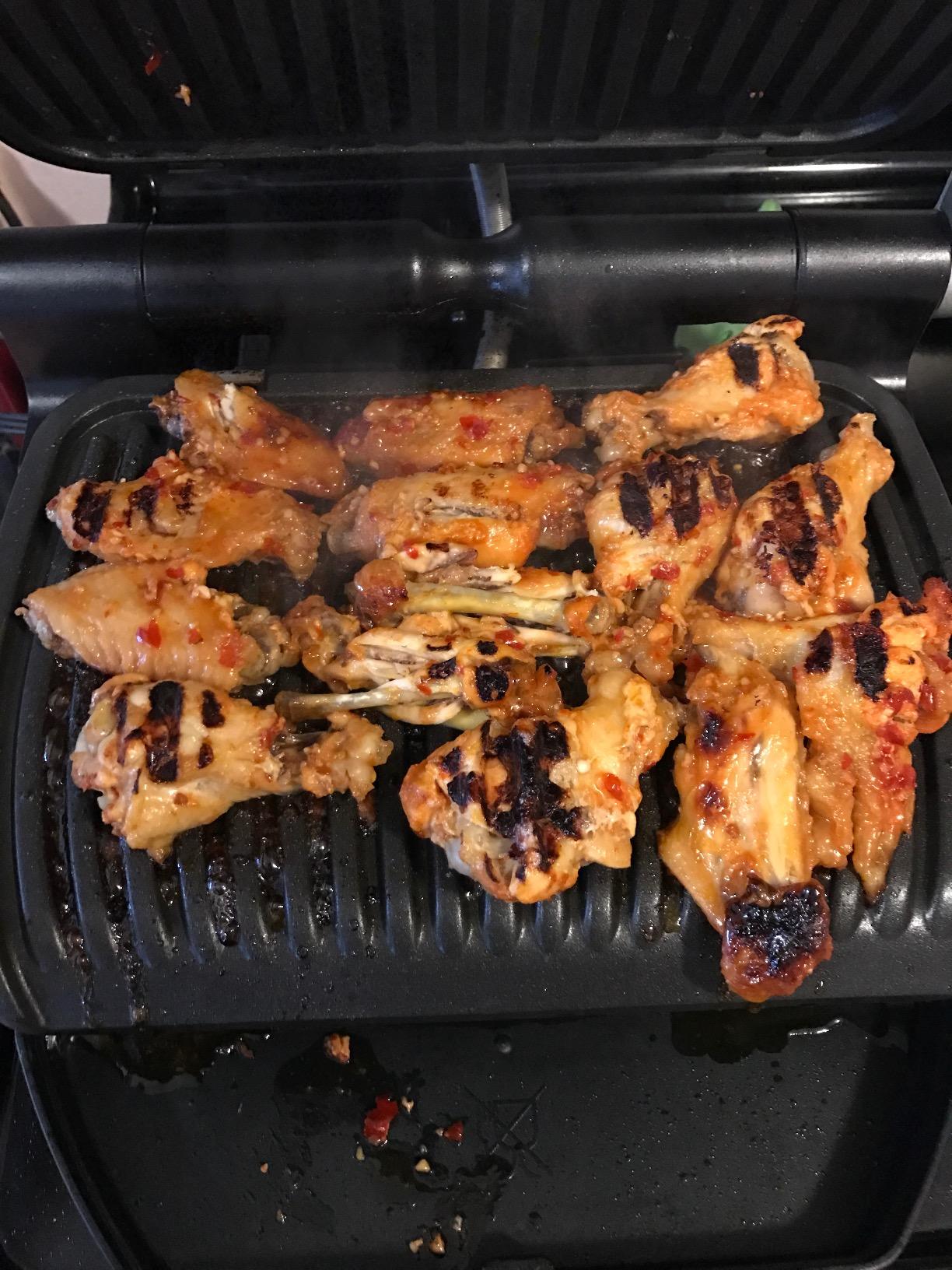}}
        \subfigure{\includegraphics[height=\csfigheight]{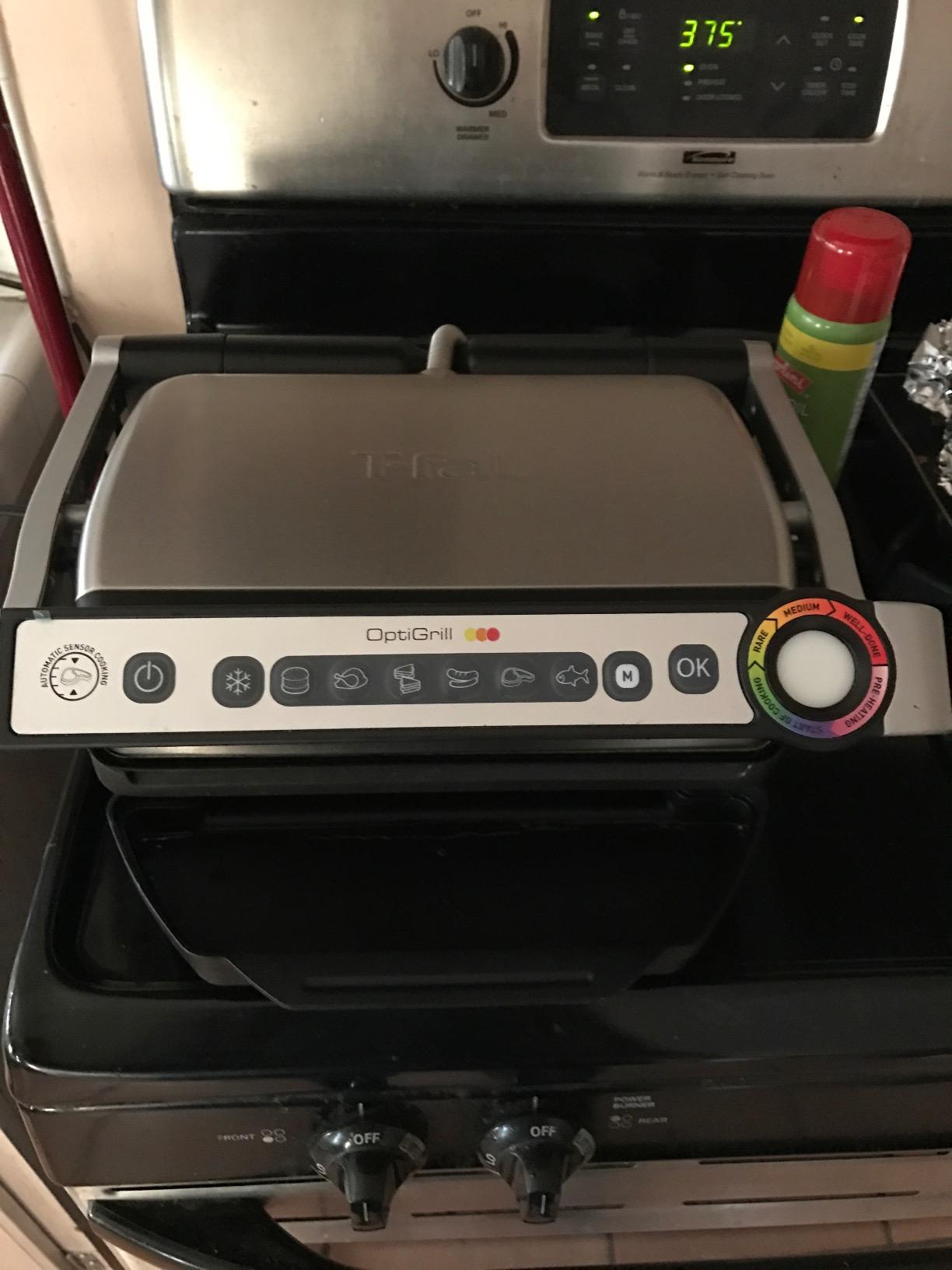}}
        \subfigure{\includegraphics[height=\csfigheight]{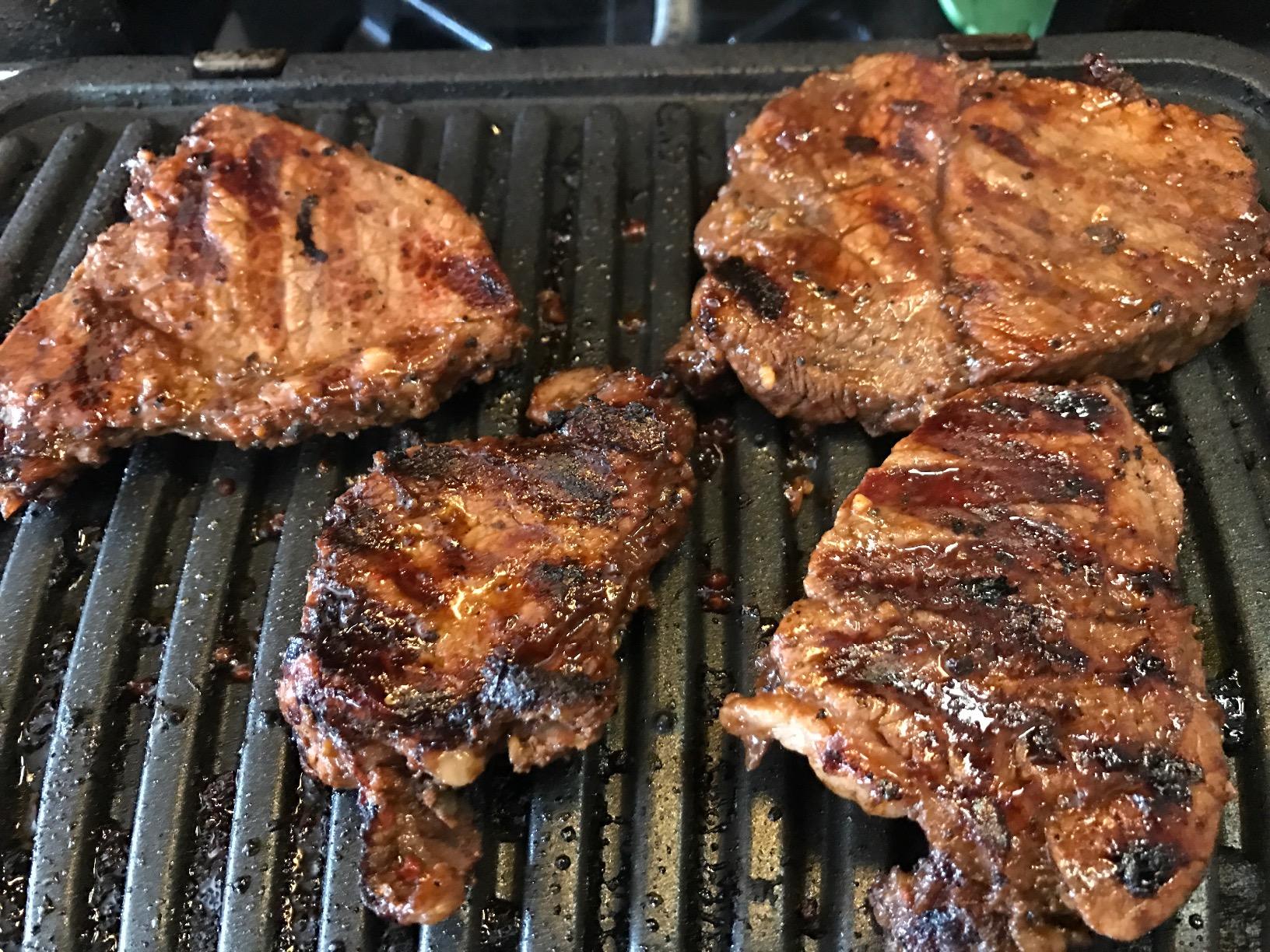}}
        \subfigure{\includegraphics[height=\csfigheight]{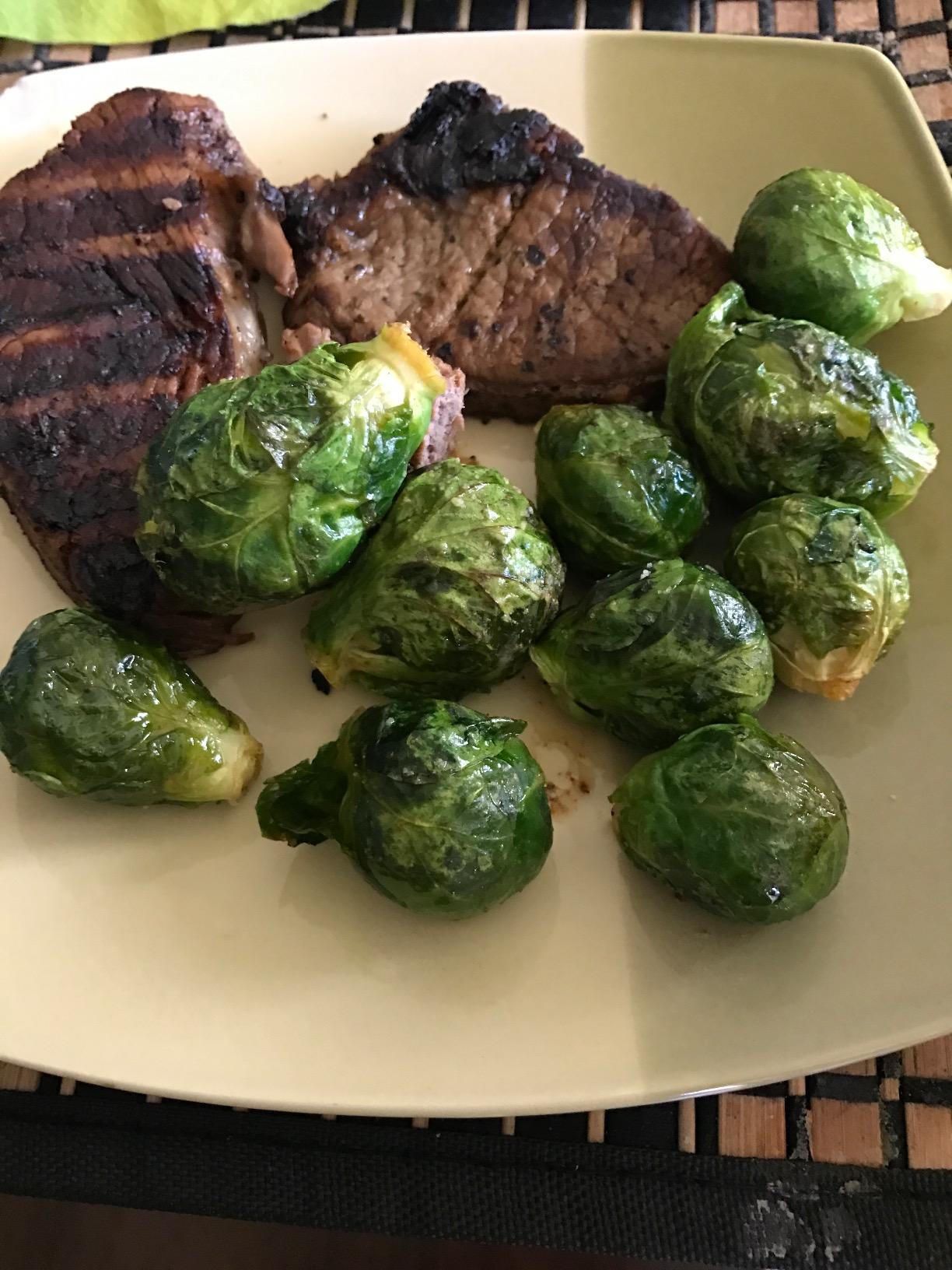}} & 1.00 & 0.35 & 0.27 \\
        \bottomrule
        \end{tabular}
    }
    \caption{(Example 2 of 2) Comparison between our model and GBDT on the example from Amazon Home dataset. The ground truth (GT) scores are annotated ones, while the scores below GBDT and ours are normalized ones.}
    \label{tab:case_study2}
\end{table*}

\end{document}